\pdfoutput=1

\documentclass[11pt]{article}
\usepackage[table,xcdraw]{xcolor}
\usepackage{multirow}
\usepackage[]{naacl2021}

\usepackage{times}
\usepackage{latexsym}

\usepackage[T1]{fontenc}

\usepackage[utf8]{inputenc}

\usepackage{microtype}

\author{Prithviraj Ammanabrolu$^{\dagger}$ \;\:\quad %
Jack Urbanek$^{\ddagger}$ \;\:\quad
Margaret Li$^{\ddagger}$ \\
\textbf{Arthur Szlam}$^{\ddagger}$ \;\;\;\:\qquad
\textbf{Tim Rockt{\"a}schel}$^{\ddagger}$ \;\;\;\:\qquad
\textbf{Jason Weston}$^{\ddagger}$\\
$^{\dagger}$Georgia Institute of Technology
\hspace{3em}
$^{\ddagger}$Facebook AI Research\\
\texttt{raj.ammanabrolu@gatech.edu} 
}

\usepackage{microtype}      %
\usepackage[linewidth=1pt]{mdframed}
\usepackage{tikz-dependency}
\usepackage{amsmath,amsfonts,bm}
\usepackage{graphicx}
\usepackage{algorithmicx}
\usepackage[noend]{algpseudocode}
\usepackage{booktabs}
\usepackage[percent]{overpic}
\usepackage{subcaption}
\usepackage{siunitx}
\usepackage{wrapfig}
\usepackage{listings}
\usepackage{multirow}
\usepackage[page]{appendix}

\usepackage{verbatim}
\usepackage{hyperref}
\usepackage{url}
\definecolor{human}{RGB}{213,235,252}
\definecolor{bot}{RGB}{224,224,224}

\NewEnviron{elaboration}{
\begin{tikzpicture}
\node[rectangle,minimum width=\textwidth,fill=bot] (m) {\begin{minipage}{.98\textwidth}\BODY\end{minipage}};
\draw[rounded corners, dotted] (m.south west) rectangle (m.north east);
\end{tikzpicture}
}
\NewEnviron{elaborationr}{
\begin{tikzpicture}
\node[rectangle,minimum width=\textwidth,fill=human] (m) {
\raggedleft
\begin{minipage}{.98\textwidth}
\BODY
\end{minipage}};
\draw[rounded corners, dotted] (m.south west) rectangle (m.north east);
\end{tikzpicture}
}

\usepackage[export]{adjustbox}
\usepackage{tabularx}
\usepackage{printlen}

\title{How to Motivate Your Dragon: \\Teaching Goal-Driven Agents to Speak and Act in Fantasy Worlds}

\begin{document}

\maketitle

\begin{abstract}

We seek to create agents that both act and communicate with other agents in pursuit of a goal.  
Towards this end, we extend LIGHT~\citep{urbanek2019light}---a large-scale crowd-sourced fantasy text-game---with a dataset of ``quests''.\footnote{Data can be found here \url{https://parl.ai/projects/light/}}.  
These contain natural language motivations %
paired with in-game goals and human demonstrations;  completing a quest might require dialogue or actions (or both). 
We introduce a reinforcement learning system that (1) incorporates large-scale language modeling-based and commonsense reasoning-based pre-training 
to imbue the agent with relevant priors; and (2) leverages a factorized action space of action commands and dialogue, balancing between the two.
We conduct zero-shot evaluations using held-out human expert demonstrations, showing that our agents are able to act consistently and talk  naturally with respect to their motivations.

\end{abstract}

\section{Introduction}

There has been a recent improvement in the quality of natural language processing (NLP) and generation (NLG) by machine learning (ML) \citep{vaswani2017attention,devlin18}; and in parallel, improvement to goal-oriented ML driven agents in the context of games \citep{vinyals2019grandmaster,schrittwieser2019mastering}.
However, agents that can communicate with humans (and other agents) through natural language in pursuit of their goals are still primitive. 
One possible reason for this is that many datasets and tasks used for NLP are static, not supporting interaction and language grounding ~\citep{brooks1991intelligence,theoryoflang,groundedcog,mikolov2016roadmap,gauthier2016paradigm,lake2017building}.
Text-based games---where players see, act upon, and communicate within a dynamic world using natural language---provide a platform on which to develop such goal-driven agents.  

LIGHT~\citep{urbanek2019light}, a large-scale crowdsourced fantasy text-adventure game, consisting of a set of locations, characters, and objectsa possesses rich textual worlds, but without any notion of goals to train goal-driven agents. %
We present a dataset of %
quests for LIGHT and demonstrations of humans playing these quests (as seen in Figures~\ref{fig:questbase}~and~\ref{fig:lightquests}), providing natural language descriptions in varying levels of abstraction of motivations for a given character in a particular setting.

To complete these quests, an agent must reason about potential actions and utterances based on incomplete descriptions of the locations, objects, and other characters.  When a human is placed in a fantasy setting such as LIGHT, they already know that kings are royalty and must be treated respectfully, swords are weapons, etc.---commonsense knowledge that a learning agent must acquire to ensure successful interactions.
To equip agents with relevant priors in such worlds, we domain-adapt the large-scale commonsense knowledge graph ATOMIC~\citep{sap2019atomic} to the LIGHT fantasy world---to build ATOMIC-LIGHT.

We then introduce a reinforcement learning (RL) system that incorporates large-scale language modeling and the above commonsense-based pre-training.   We show that RL is superior to behavior cloning or other supervised training on our data; and that carefully combining pre-training with RL is superior to either.

\begin{figure*}[t!]
    \centering
    \scriptsize
\begin{minipage}{\textwidth}
\begin{tabularx}{\textwidth}{|r|X|}
\hline
\multicolumn{1}{|l|}{\textbf{Setting}}   & You are in the Dangerous Precipice. The dangerous precipice overlooks the valley below.  The ground slopes down to the edge here.  Dirt crumbles down to the edge of the cliff. There's a dragon crescent, a knight's armor, a golden dragon egg, and a knight's fighting gear here. A knight is here.You are carrying nothing.  \\ \hline
\rowcolor[HTML]{E0E0E0}
\multicolumn{1}{|l|}{\cellcolor[HTML]{E0E0E0}{\textbf{Partner:}}}   & Knight.   \\
\rowcolor[HTML]{E0E0E0}
Persona & I am a knight. I come from a lower-ranking noble family. I serve under the king, as my father did before me. In times of war, I fight on horseback.  \\
\rowcolor[HTML]{E0E0E0}
Carrying & knight's armor, golden dragon egg, knight's fighting gear \\ \hline
\rowcolor[HTML]{D5EBFC}
\multicolumn{1}{|l|}{\cellcolor[HTML]{D5EBFC}{\textbf{Self:}}} &  A dragon.\\
\rowcolor[HTML]{D5EBFC}
Persona & I am a dragon living in the mountains. I enjoy hoarding treasure. I terrorize the local populace for fun.  \\
\rowcolor[HTML]{D5EBFC}
Carrying & Nothing.  \\
\hline
\end{tabularx}
\vspace{-8pt}
\caption{Setting and character information for both self and partner characters as taken from LIGHT.}
\label{fig:lightbase}
\begin{tabularx}{\textwidth}{|
>{\columncolor[HTML]{D5EBFC}}r| 
>{\columncolor[HTML]{D5EBFC}}l| 
>{\columncolor[HTML]{D5EBFC}}r|
>{\columncolor[HTML]{D5EBFC}}X| }
\hline
\multicolumn{2}{|l}{\cellcolor[HTML]{D5EBFC}\textbf{Motivations:}}                                                                                                            & \multicolumn{2}{|l|}{\cellcolor[HTML]{D5EBFC}\textbf{Timeline:}} \\ \hline
\cellcolor[HTML]{D5EBFC}                        & \cellcolor[HTML]{D5EBFC}                                                                                          & -4 hours        & go to dangerous precipice           \\
\multirow{-2}{*}{\cellcolor[HTML]{D5EBFC}Short} & \multirow{-2}{*}{\cellcolor[HTML]{D5EBFC}I need to recover the dragon egg that was stolen and punish the knight.} & -15 min         & get knights armor from knight       \\
\cellcolor[HTML]{D5EBFC}                        & \cellcolor[HTML]{D5EBFC}                                                                                          & -10 min         & get golden dragon egg               \\
\cellcolor[HTML]{D5EBFC}                        & \cellcolor[HTML]{D5EBFC}                                                                                          & Now             & hit knight                          \\
\multirow{-3}{*}{\cellcolor[HTML]{D5EBFC}Mid}   & \multirow{-3}{*}{\cellcolor[HTML]{D5EBFC}I need to return the golden dragon egg to my treasure hoard.}      & +5 min          & put dragon egg on back              \\
\cellcolor[HTML]{D5EBFC}                        & \cellcolor[HTML]{D5EBFC}                                                                                          & +15 min         & eat the knight                      \\
\multirow{-2}{*}{\cellcolor[HTML]{D5EBFC}Long}  & \multirow{-2}{*}{\cellcolor[HTML]{D5EBFC}I need to build the largest hoard ever attained by any one dragon.}            & +2 hours        & go to the mountains       \\ \hline         
\end{tabularx}
\vspace{-8pt}
\caption{Motivations with different levels of abstractions and corresponding sequence of timeline actions in chronological order for the self character in LIGHT-Quests. There are $7486$ quests in total.}
\label{fig:questbase}
\end{minipage}
\begin{elaborationr}
  \parbox{.99\textwidth}{
\hspace*{\fill} Insssssolent pessst!  I should immolate you for this tresssspasss. 
}
\end{elaborationr}
\begin{elaboration}
  \parbox{.99\textwidth}{
And why is that, dragon?
}
\end{elaboration}
\begin{elaborationr}
  \parbox{.99\textwidth}{
\hspace*{\fill} Ssstealing my preccciousss golden egg!
I'll tell you what, I'll give you 10 sssseconds to amussse me with your sssstory and THEN I'll burn you alive! 
}
\end{elaborationr}
\begin{elaboration}
  \parbox{.99\textwidth}{
You said you wanted to attack me, dragon, did you not?
}
\end{elaboration}
\begin{elaborationr}
  \parbox{.99\textwidth}{
\hspace*{\fill} Go ahead, I'm lisssssstening.\\
\hspace*{\fill} {\em get golden dragon egg}
}
\end{elaborationr}
\begin{elaboration}
  \parbox{.99\textwidth}{
Now now! I would have given you that had you asked!
}
\end{elaboration}
\begin{elaborationr}
  \parbox{.99\textwidth}{
\hspace*{\fill} Assssssk for my own property back? What a riduculousss notion
}
\end{elaborationr}
\begin{elaboration}
  \parbox{.99\textwidth}{
Look here, I told you to watch your mouth and you didn't, so leave or I'll make you leave.}
\end{elaboration}
\begin{elaborationr}
  \parbox{.99\textwidth}{
\hspace*{\fill} And now threatsss!  Thisss is proving to be a mossst engaging conversssation. \\
\hspace*{\fill} {\em hit knight} \\
\hspace*{\fill} Give my regardsss to the valley floor below!
}
\end{elaborationr}
\vspace{-15pt}
\caption{Example of a demonstration of a human
(blue shaded) %
completing the above quest while role-playing as the self character with a partner agent (grey shaded).
There are $2111$ such human demonstrations of average sequence length $12.92$,
consisting of $22672$ dialogues in total.}
\label{fig:lightquests}
\end{figure*}

However, we find that although pre-training can be an effective tool in this setting, it requires more finesse than in the standard supervised setting.   In particular, we find that simply pre-training a model on a large ``generic'' corpus~\citep{sap2019atomic,baumgartner2020pushshift} of commonsense/language data or pre-training on the domain specific LIGHT corpus, and then fine-tuning via RL is {\it less} effective than training RL from scratch. Furthermore, by carefully combining general and domain-specific pre-training, we observe large improvements over RL from scratch.

In short, the contributions of this paper are threefold: (1) A dataset of quests, LIGHT-Quests, 
and a companion fantasy themed commonsense knowledge graph ATOMIC-LIGHT; (2) a reinforcement learning architecture and training methodology that use these datasets to create goal-driven agents that act and speak in the LIGHT environment; and (3)
Empirical zero-shot evaluations based on human quest demonstrations and an analysis of large-scale transformer-based pre-training trends in static vs. interactive settings, showing that we have trained agents that act consistently and speak 
naturally with respect to their motivations.

\section{Related Work}

We focus on four major areas of related work: text-based game-playing, goal-oriented dialogue, commonsense reasoning in language, and general language-informed RL.

\textbf{Text-based game-playing.}
\cite{cote18} introduce TextWorld, a framework for procedurally generating text-based games via grammars, and \cite{yuan18,jonmay,adolphs2019ledeepchef,adhikari2020learning} build agents that operate in this environment---focusing on aspects such as efficient exploration and zero-shot generalization to new, procedurally generated environments.
Similarly, \cite{jericho} introduce Jericho, a framework and series of baseline agents for interacting with human-made text-games such as {\em Zork}~\citep{zork}.
This resulted in agents developed by works such as \cite{zahavy18,Ammanabrolu2020Graph}, aiming to learn to execute contextually relevant actions.
Other works such as \cite{narasimhan15,he16} explore how to best factorize such text-game action spaces.
None of these works consider agents with motivations and personas nor require any dialogue.

\textbf{Goal-oriented dialogue.}
This form of dialogue has traditionally been closely related to specific tasks useful in the context of personal assistants with dialogue interfaces~\citep{henderson2014second,asri2017frames}.
RL has been studied for such tasks, usually to improve dialogue state management~\citep{singh2000reinforcement,pietquin2011sample,fatemi2016policy} and to improve response quality~\citep{li16deep}.
In particular, the negotiation tasks of \cite{yarats2017hierarchical,lewis2017deal}, where two agents are trying to convince each other to perform certain actions, are related to the tasks in LIGHT-Quests.
These works all lack environment grounding 
and the notion of diverse agent motivations.

\textbf{Commonsense reasoning in language.}
Works such as \cite{Bosselut2019COMETCT,guan2020knowledge} focus on pre-training transformer-based 
language learning systems with large-scale commonsense knowledge graphs such as ATOMIC~\citep{sap2019atomic} and ConceptNet~\citep{Speer2012} for use in knowledge graph completion and story ending generation respectively.
\cite{fulda17,ammanabrolutransfer,ammanabrolu2020avoid,murugesan2020enhancing} look at commonsense reasoning in interactive environments, with the former focusing on affordance extraction using word embeddings and the latter three on transferring text-game playing skills via pre-training using question-answering and large-scale knowledge graphs. %

\textbf{Language-informed reinforcement learning.}
\cite{luketina2019survey} provide an overview of RL informed by natural language.
Of these works, the ones most related to ours are those falling into the category of instruction following---where an agent's tasks are defined by high level instructions describing desired policies and goals~\citep{macmahon2006walk,kollar2010toward}.
Visual and embodied agents using natural language instructions~\citep{bisk2016natural,ai2thor,anderson2018vision} or in language-based action spaces~\citep{das2017learning} utilize interactivity and environment grounding but have no notion of agent motivations, nor make any attempt to explicitly model commonsense reasoning.
Perhaps closest in spirit to this work is \cite{prabhumoye2020love}, where they use artificially selected goals in LIGHT and train RL agents to achieve them.
Similarly to the others, this work does not contain the motivations provided by LIGHT-Quests nor any modeling of commonsense reasoning.
Further, they limit their RL problem to  1 and 3-step trajectories that only involve speech, and no actions---compared to the human demonstrations in LIGHT-Quests which contain both actions and speech sequences of average length $12.92$.

\section{LIGHT-Quests and ATOMIC-LIGHT}
\label{sec:datasets}
This section first provides a brief overview of the LIGHT game environment, followed by descriptions of the LIGHT-Quests and ATOMIC-LIGHT datasets used in this paper.

\textbf{Background.} The LIGHT game environment is a multi-user fantasy text-adventure game consisting of a rich, diverse set of characters, locations, and objects (1775 characters, 663 locations,  and 3462 objects).
Characters are able to perform templated actions to interact with both objects and characters, and can speak to other characters through free form text.
Actions in text games generally consist of verb phrases (VP) followed optionally by prepositional phrases (VP PP).
For example, {\em get OBJ, put OBJ, give OBJ to CHAR}, etc..
There are $13$ types of allowed verbs in LIGHT.
These actions change the state of the world which is expressed to the player in the form of text descriptions.

\subsection{LIGHT-Quests}
\label{sec:lightquests}
Figures~\ref{fig:lightbase},~\ref{fig:questbase}, and~\ref{fig:lightquests} summarize the data that we collected for LIGHT-Quests.
Data is collected via crowdsourcing in two phases, first the quests then demonstration of humans playing them.
During the first phase, crowdworkers were given a setting, i.e. situated in a world, in addition to a character and its corresponding persona and asked to describe in free form text what potential motivations or goals could be for that character in the given world.
The kind of information given to the crowdworkers is seen in Figure~\ref{fig:lightbase}.
Simultaneously, they were also asked to provide a sequence of seven timeline actions---one action that needs to be completed {\em now} and three before and after at various user-defined intervals---for how the character might go about achieving these motivations. %

Given the information in Figure~\ref{fig:lightbase}, the crowdworkers completed the above outlined tasks and produce data as seen in Figure~\ref{fig:questbase}.
Motivations come in three levels of abstraction---short, mid, and long---corresponding to differing amounts of the timeline.
For example, the short motivation is always guaranteed to correspond most closely to the {\em now} position on the timeline.
Action annotation is pre-constrained based on the classes of verbs available within LIGHT.
The rest of the action is completed as free form text as it may contain novel entities introduced in the motivations.
There are $5982$ training, $756$ validation, and $748$ test quests.
Further details regarding the exact data collection process and details of LIGHT-Quests are found in Appendix~\ref{app:questmturk}.

After collecting motivation and timelines for the quests, we deployed a two-player version of the LIGHT game, 
letting players attempt the quests for themselves in order to collect human demonstrations.
Figure~\ref{fig:lightquests} shows an example human expert demonstration of a quest.
Players were given a character, setting, motivation, and a partner agent and left to freely act in the world and talk to the partner in pursuit of their motivations.
The partner agent is a fixed poly-encoder transformer model \citep{Humeau2020Poly-encoders:} trained on the original LIGHT data as well as other human interactions derived via the deployed game---using 111k utterances in total. %
Players first receive a role-playing score on a scale of 1-5 through a Dungeon Master (DM), a learned model that ranks how likely their utterances are given the current context.
Once they have accumulated a score reaching a certain threshold, they are allowed to perform actions.
We employ this gamification mechanism to encourage 
players to role-play their character persona and its motivations, leading to improved user experience and data quality \citep{horsfall2011study}.
They are then given further reward if the actions they perform sequentially match those on the timeline for the given quest.
The game ends after a maximum of six turns of dialogue per agent, i.e. twelve in total.
The average sequence of a human demonstration is $12.92$, with an average action sequence length of $2.18$ and dialogue of $10.74$.
There are $1800$ training, $100$ validation, and $211$ test human expert demonstrations after the data was filtered.
Additional details and examples are found in Appendix~\ref{app:lightquestdemos}.

\subsection{ATOMIC-LIGHT}
\label{sec:atomiclight}
Commonsense reasoning is a critical cornerstone when building learning agents that navigate spaces such as LIGHT-Quests.
To this end, we domain-adapt the large-scale commonsense knowledge base ATOMIC~\citep{sap2019atomic} to LIGHT.
ATOMIC contains information relevant for everyday commonsense reasoning in the form of typed if-then relations with variables.
ATOMIC is organized into a set of events, e.g. ``X puts X's trust in Y'' and annotated relation types such as ``needs'', ``wants'', ``attributes'', and ``effects'' that label the effects.
It is designed to be a general atlas of commonsense data and so is neither dependent on a specific environment or a character's persona and motivations.

To construct ATOMIC-LIGHT, we specifically use the relations for ``intents'', ``effects'', ``wants'' and ''needs'' and expand the $\langle subject,relation,object\rangle$ triples found in the graph into templated natural language sentences.
These sentences are then rewritten to better reflect the fantasy LIGHT domain.
Named entities and other noun phrases in ATOMIC are masked out and filled in using  BERT~\citep{devlin18} fine-tuned using a masked language model loss on the entire LIGHT and LIGHT-Quests data.
We investigate the benefits of such domain adaptation on downstream tasks in Section~\ref{sec:pretrainingtasks}.
An example of a clause using the {\em wants} relation in ATOMIC is as follows,
``{\em PersonX puts PersonX trust in PersonY}, {\em wants}, {\em rely on PersonY}.''
In ATOMIC-LIGHT, this is rewritten to: 
``\underline{The merchant puts the merchant's trust in the} \underline{guard}, as a result the merchant {\em wants} to \underline{rely on the guard}.''
Similarly, an example of an effect using the {\em needs} relation is, ``Before, \underline{the merchant puts the merchant's  trust in the} \underline{guard}, the merchant {\em needs} to \underline{be friends} \underline{with the guard}.''
ATOMIC-LIGHT contains $216686$ training, $35340$ validation, and $38565$ test samples.
Further details of the construction of this dataset are found in Appendix~\ref{app:atomiclight}.

\section{Agents that Act and Speak}
\label{sec:lightrl}
This section describes the creation of the agents that learn to act and speak conditioned on their motivations in the LIGHT environment.
The overall architecture and training are first outlined, followed by a detailed discussion on types of encoder pre-training.

\begin{figure*}
    \centering
    \includegraphics[width=0.85\linewidth]{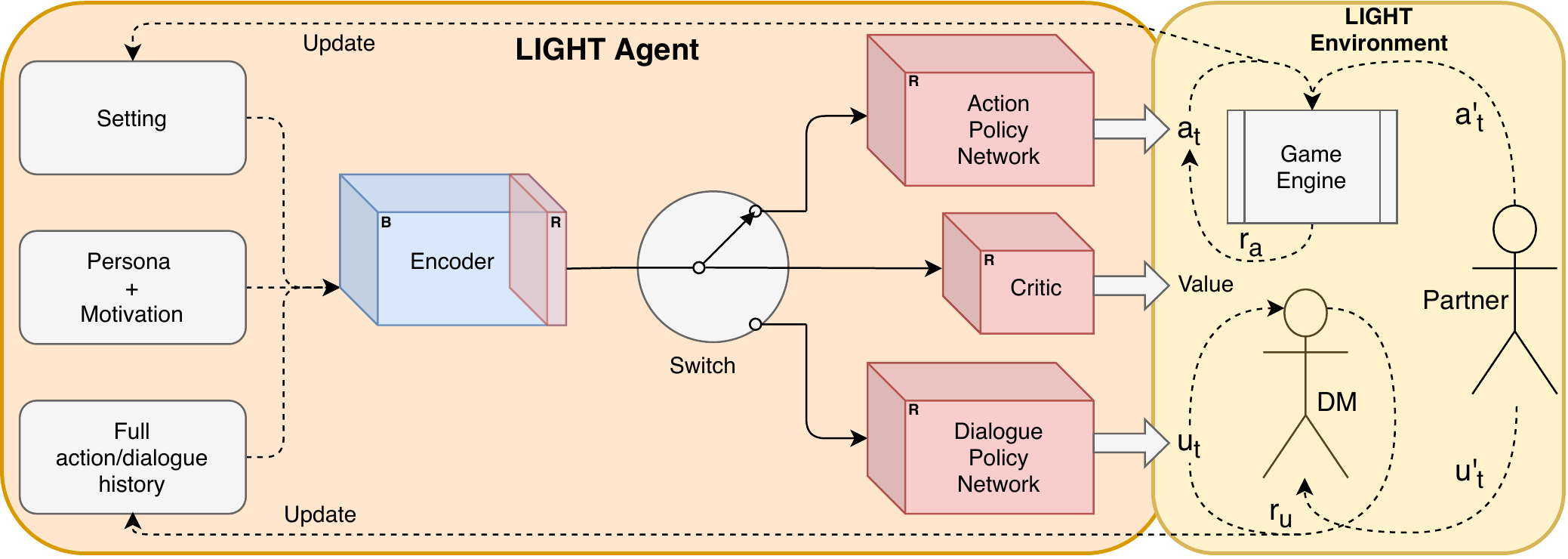}
    \caption{Overall RL Switch architecture and process. \textbf{B}lue shaded components can be pre-trained and \textbf{R}ed shaded components are trained with RL. Solid lines indicate gradient flow.}
    \label{fig:rlswitch}
\end{figure*}

\subsection{LIGHT RL Environment}
\label{sec:env}
The environment as seen in Figure~\ref{fig:rlswitch} consists of three components.  
The first is a partner agent, which is a model trained to play other agents in the game, as in \cite{prabhumoye2020love}. 
Next is the game engine, which determines the effects of actions on the underlying game graph \citep{urbanek2019light}.    
Finally, there is the Dungeon Master (DM), which is trained to score the naturalness of dialogue. 

\textbf{Partner Agent.}
The partner agent is a poly-encoder transformer model \citep{Humeau2020Poly-encoders:} that is pre-trained on the Reddit dialogue corpus, then on LIGHT and the human demonstrations of LIGHT-Quests.
Following the format seen in Figure~\ref{fig:lightquests}, the partner agent does not have a motivation itself but is trained to react to agents with motivations.
Following~\cite{prabhumoye2020love}, we keep the partner model fixed during the episodes where the LIGHT agent trains to ensure that it retains natural English semantics---avoiding the problem of language drift by learning an emergent language with that must agree with the partner's usage~\citep{lee2019countering}.

\textbf{Action Rewards via the Game Engine.}
All actions, either those of the agent-in-training or the partner agent, are processed by the engine, checking for goal state completion---hence known as {\em act goals}.
For example, if the LIGHT agent had the motivation to acquire a sword, the goal could be completed via a:
\begin{enumerate}
    \item {\bf self act completion}: where the agent acquires a sword itself by picking it up, stealing it, convincing the partner to drop theirs so you can pick it up, etc.
    \item {\bf partner act completion}: where the agent uses speech to convince their partner to achieve the goal for them (e.g., by persuading the partner to give them the sword).
    
\end{enumerate}
Reaching an {\em act goal} provides reward $r_a$ of $1$ and $0$ otherwise.
At each step, the engine also provides us with the set of valid actions. %
These are the subset of the action space $A$ which are guaranteed to be a valid change to the world from the current state $s_t$, i.e. an action to give your partner a sword cannot be valid unless you possess the sword.

\textbf{Speech Rewards via the Dungeon Master.}
Following prior works on using transformers for automatic evaluation of natural language generation~\citep{sellam-etal-2020-bleurt}, we utilize a learned model--the Dungeon Master (DM)---to score the agent's ability to speak.
The DM used here is a poly-encoder model trained on collected human quest demonstrations %
as well as the original conversations in LIGHT. 
It is conditioned on quests and motivations and thus able to provide a (noisy) indication of how natural the agent's dialogue utterances are given its immediate context, similarly to the function of the DM during the data collection process.
Given the dialogue portion of a human quest demonstration %
of length $n$, the DM returns a reward $r_u$ of $\frac{1}{2n}$ if
an utterance was in the demonstration
(for a maximum of one time per episode for each utterance from the demonstration). %
A further $\frac{1}{2n}$ is given each time the utterance is scored as being within the top-$k$ most likely utterances by the DM.
This naturalness objective will be hence referred to as a {\em speech goal}.
These rewards thus also denser than {\em act goals}, helping the agent learn overall.
Further, similarly to the game engine, the DM also provides a set of $M$ valid utterances
which are the $M$ most likely dialogue candidates from the candidate set for the current context.

\subsection{Training a LIGHT agent with Switch Reinforcement Learning}

The overall architecture of our agent is shown in Figure~\ref{fig:rlswitch}.  It consists of an encoder, a switch, an action network, and a dialogue network. 
First, we construct the action spaces---factorized into actions and utterances.  The possible actions are the set of all actions taken in the demonstrations (4710 total) and the possible utterances are all utterances from the demonstrations (22672 total).
The encoder network processes the setting, persona, motivation, as well as the full history of actions and dialogues performed by the agent and the partner, input as a text sequence.
The features from the encoder, which here are the hidden states at the final layer of a transformer, are used as input by all following components of the agent.   
In Section~\ref{sec:eval} we show how different encoder training data affects the model. 

Next, a switch module  makes the decision regarding whether the agent should act or talk in the current context and activates the corresponding policy network.   
In this work, the switch is simple: it outputs an action every $k$ dialogue utterances; where during training $k$ is chosen to match the ratio of utterances to actions on that particular quest from the human demonstrations, and during testing, $k$ is chosen to match the average action to utterance ratio.
Both the action and dialogue  policies consist of a a single GRU layer followed by an $n$-layer feed-forward network given input features from the encoder.  
Once the LIGHT agent has output an utterance or action, it is processed by the environment---the partner agent,  the game engine and the DM.

We use A2C~\citep{mnih2016asynchronous} to train the LIGHT agent, treating the two policy networks as two separate actors with a shared critic.   The shared critic is motivated by the concepts of {\em self act completion} and {\em partner act completion} seen in Section~\ref{sec:env} where the LIGHT agent can speak to convince the partner to achieve an {\em act goal}.
Each agent in a batch is initialized via priority sampling \citep{graves2017automated} with a different quest, i.e. quests that the agent has historically successfully completed less often are given a greater weight when sampling from the pool of all possible training quests.
In addition to a normal entropy regularization term, we also add a regularization term that encourages the models to produce ``valid'' outputs as judged by the game engine and the DM for actions and utterances respectively.
Additional training details are found in Appendix~\ref{app:lightrl}.

\subsection{Encoder Pre-training Tasks}
\label{sec:pretrainingtasks}

Prior work on commonsense reasoning in supervised natural language learning~\citep{Bosselut2019COMETCT} suggests that the encoder is key to overcoming the challenges posed by the LIGHT-Quests dataset even in an RL setting.
We describe a series of encoder pre-training tasks, designed to help the LIGHT agent either act more consistently or speak more naturally.

\textbf{ATOMIC-LIGHT} 
As seen in Section~\ref{sec:datasets}, ATOMIC-LIGHT is a (domain-adapted) fantasy commonsense knowledge graph, and as such provides priors for an agent on how to act consistently in the world.
For example, given a clause such as
``The knight wishes to slay the dragon, as a result the knight {\em needs} to \underline{acquire a sword},'' the task would be to predict the underlined text---a form of knowledge graph completion~\citep{kgcompletion}.

\textbf{Reddit} We use a previously existing Reddit dataset extracted and obtained by a third party and made available on pushshift.io ~\citep{baumgartner2020pushshift} seen in \cite{roller2020recipes}.
This dataset has been used in several existing dialogue-based studies and has been shown to result in more natural conversations~\citep{yang2018learning,mazare-etal-2018-training}.

\textbf{LIGHT-Original} 
The original LIGHT dataset~\citep{urbanek2019light} is organized similarly to the human demonstrations found in LIGHT-Quests, i.e. an interspersed sequence of dialogue and actions collected from humans role-playing a character.
The task itself is to predict the next action or utterance given the prior dialogue history as well as the current setting and persona for a character.
They are collected in a chit-chat fashion, with no notion of objectives, and so provide priors on how to generally act consistently and speak in a fantasy world, but not directly how to complete quests.

\textbf{LIGHT-Quests}
Pre-training with this newly introduced
dataset consists of three tasks.
(1) {\em Bag-of-action timeline prediction} in which, given a quest consisting of setting, persona, and motivations, any one of the actions in the timeline must be predicted.
(2) {\em Sequential timeline prediction} in which, given a quest consisting of setting, persona, motivations, and the first $n$ actions in the timeline, the $n+1^{th}$ action must be predicted.
(3) Predict the next dialogue utterance given a human demonstration in a manner similar to the LIGHT-original tasks.
The first two tasks are designed to help the agent act consistently and the third to help it speak naturally with respect to its motivations.

\begin{table*}[]
\centering
\footnotesize
\begin{tabular}{|l|l|l|l|l|}
\hline
\multicolumn{1}{|c|}{\multirow{2}{*}{\textbf{Model}}} & \multicolumn{3}{c|}{\textbf{Reinforcement Learning}}                                                                                     & \multicolumn{1}{c|}{\textbf{Behavioral Cloning}}    \\ 
\multicolumn{1}{|c|}{}                                & \multicolumn{1}{c|}{\textbf{Act Goals}} & \multicolumn{1}{c|}{\textbf{Speech Goals}} & \multicolumn{1}{c|}{\textbf{Act \& Speech Goals}} & \multicolumn{1}{c|}{\textbf{Act \& Speech Goals}} \\ \hline
\textbf{Scratch}                                               & 0.418                           & 0.118                               & 0.103                                    & 0.0003                                     \\
\textbf{General}                                                & 0.146                            & 0.040                               & 0.028                                      & 0.00226                                    \\ %
\textbf{Light}                                          & 0.115                             & 0.028                               & 0.022                                      & 0.0934                                    \\ %
\textbf{General+Light}                                      & 0.251                             & 0.094                                 & 0.081                                       & 0.115                                       \\ %
\textbf{Adaptive}                                                  & \textbf{0.420}                             & \textbf{0.330}                                & \textbf{0.303}                                      & \textbf{0.147}                                      \\
\hline   
\end{tabular}
\caption{Encoder Type RL Zero-Shot Evaluations averaged over 3 independent runs. Act goals and speech goals are as described in Section~\ref{sec:env}. Standard deviations for all experiments are less than 0.01. The ``Act \& Speech Goals'' column refers to quests where the agent has simultaneously achieved both types of goals within the episode. Human $\text{act goal completion}=0.6$ as measured during the second phase of the LIGHT-Quests data collection.}
\label{tab:enctyperes}
\end{table*}

\section{Evaluation}
\label{sec:eval}
We conduct two ablation studies, (1) to compare the effects of the encoder pre-training tasks in RL settings vs. supervised behavior cloning, and (2) to analyze the interplay between actions and dialogue for {\em self} and {\em partner act completions}.

\subsection{Encoder Pre-training Type Ablation Study}
Pre-training is done on the tasks described in Section~\ref{sec:pretrainingtasks}  by  training a 12 layer transformer with $256$ million parameters using a cross-entropy loss as seen in~\cite{Humeau2020Poly-encoders:}.
These weights are then transferred to the \textbf{B}lue shaded portion of the encoder as seen in Figure~\ref{fig:rlswitch} and frozen.
A further three randomly initialized-layers are appended on to the end, indicated by the \textbf{R}ed portions, into which gradients flow.
This is done as optimizing all the parameters of such a model via RL over a long horizon is both data inefficient and computationally infeasible.
Additional hyperparameter details are found in Appendix~\ref{app:lightquestssup}.
We investigate the following five different pre-training models to see how they compare on {\em act} and {\em speech goal completions} when trained with RL and in a supervised manner with behavior cloning:

\textbf{Scratch} No pre-training is done, the encoder is a 3-layer randomly initialized transformer and trained along with the policy networks.
\begin{figure*}
    \centering
    \begin{subfigure}{0.39\textwidth}
    \centering
    \includegraphics[width=\linewidth]{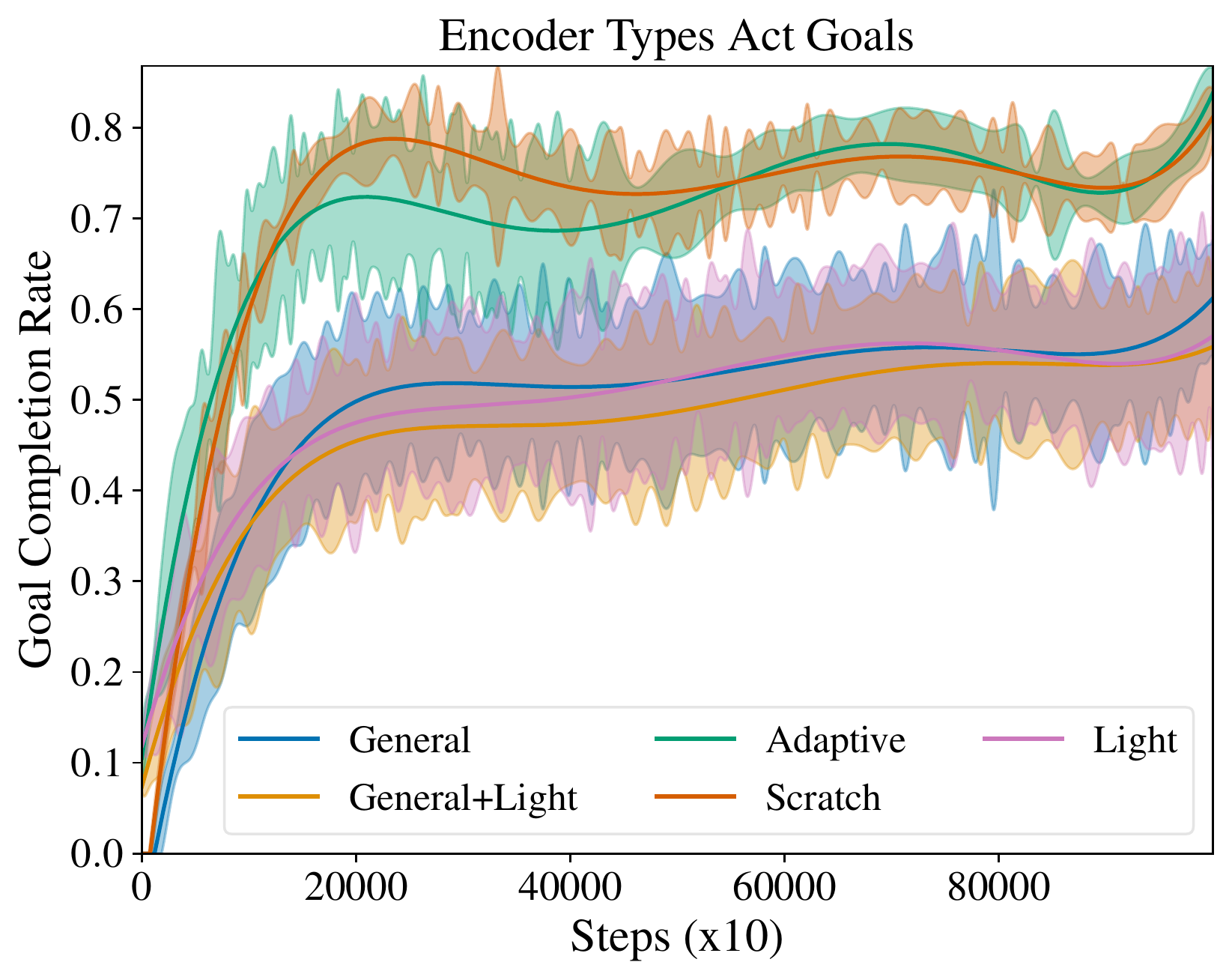}
    \end{subfigure}
    \begin{subfigure}{0.39\textwidth}
    \centering
    \includegraphics[width=\linewidth]{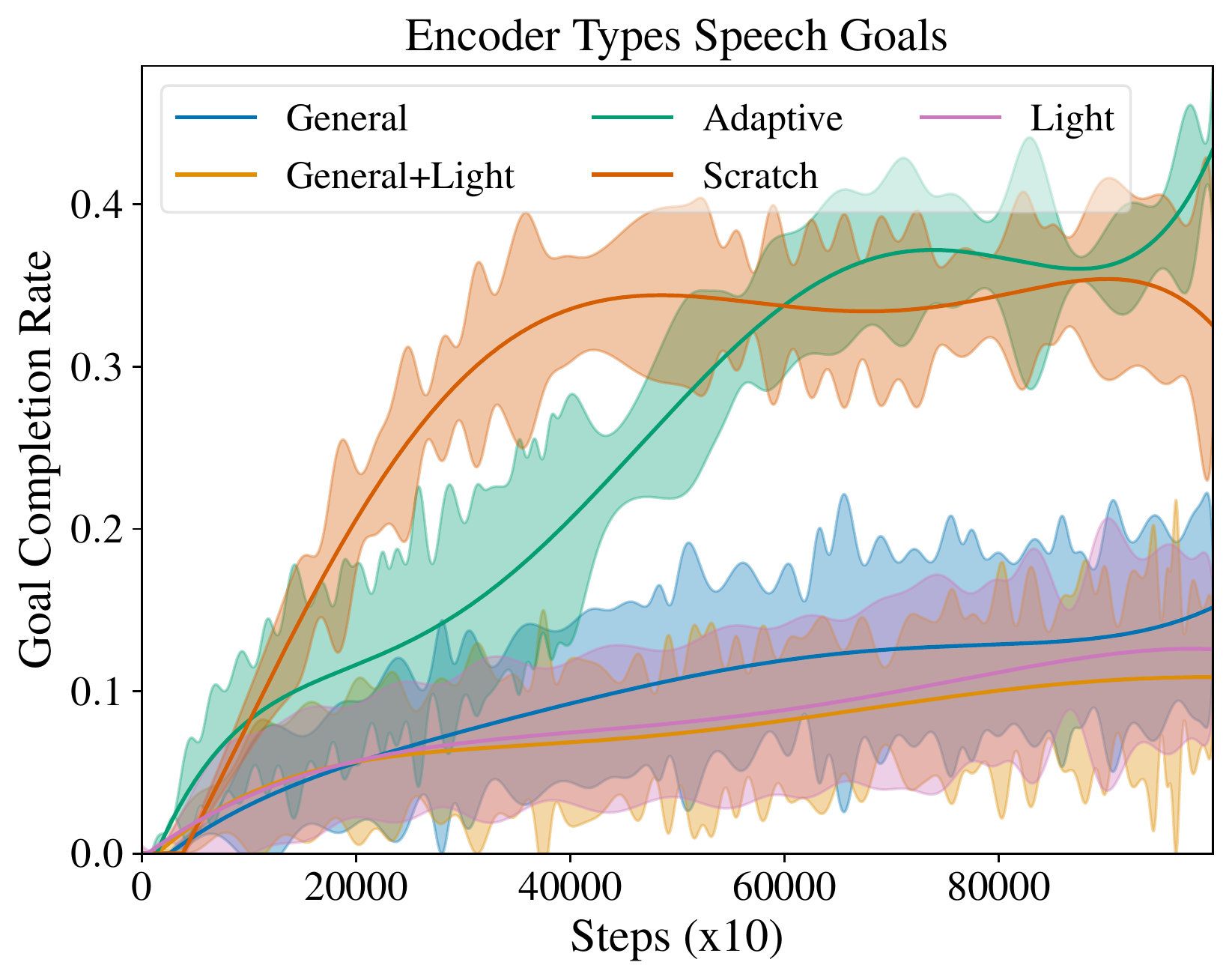}
    \end{subfigure}
    \caption{Encoder types RL reward curves averaged over 3 independent runs.}
    \label{fig:encoders}
\end{figure*}

\textbf{General} Multi-task trained using both pushshift.io Reddit and the commonsense dataset ATOMIC-LIGHT, giving the agent general priors on how to act and speak.

\textbf{Light} Multi-task trained on all tasks in LIGHT-original and LIGHT-Quests, giving the agent priors on how to act and speak with motivations in the LIGHT fantasy domain.

\textbf{General+Light} Multi-task trained on all tasks used in the General and Light models.

\textbf{Adaptive} Here we adaptively train a General+Light model that is first initialized itself from a General model, providing additional regularization to help balance between Light and General tasks.

Table~\ref{tab:enctyperes} describes the results for this ablation.
Models were each zero-shot evaluated on $211$ human demonstrations from the LIGHT-Quests test set for a single episode per quest across three independent runs.
Figure~\ref{fig:encoders} shows learning curves during training for each encoder type.
We first see that performance when trained with RL, i.e. with interactivity and environment grounding during training, results in higher performance than behavioral cloning for all the models.
In both RL and behavior cloning settings the Adaptive model outperforms all others in all the metrics.%

When trained supervised (behavioral cloning), we see trends mirroring standard pre-training in static text corpora.
Transfer is easy and the Scratch model performs significantly worse than all others; and each new task added improves the agent's ability to speak and act.
In particular, we see that Light outperforms General, showing that the more similar the pre-training tasks are to the downstream tasks, the better the supervised performance.

\begin{table*}[!h]
\centering
\footnotesize
\begin{tabular}{|l|l|l|l|l|l|l|l|}
\hline
\multicolumn{1}{|c|}{\multirow{2}{*}{\textbf{Ability}}} & \multicolumn{3}{c|}{\textbf{Scratch}}                                                                                              & \multicolumn{3}{c|}{\textbf{Adaptive}}                                                                                                 \\
\multicolumn{1}{|c|}{}                                  & \multicolumn{1}{c|}{\textbf{Act Goals}} & \multicolumn{1}{c|}{\textbf{Speech Goals}} & \multicolumn{1}{c|}{\textbf{Act \& Speech}} & \multicolumn{1}{c|}{\textbf{Act Goals}} & \multicolumn{1}{c|}{\textbf{Speech Goals}} & \multicolumn{1}{c|}{\textbf{Act \& Speech}} \\ \hline

Act+Speech & 0.418                           & 0.118                               & \textbf{0.103}                                     & 0.420                             & 0.330                                & \textbf{0.303}                                                                            \\
Act Only                               & \textbf{0.478}                                   & -                                          & -                                           & \textbf{0.469}                                   & -                                          & -                                           \\
Speech Only                            & 0.036                                   & \textbf{0.165}                                      & 0.028                                       & 0.0398                                  & \textbf{0.341}                                      & 0.030                                       \\
\multicolumn{1}{|r|}{-No Speech Goals} & 0.0526                                  & 0.0521                                     & 0.0331                                      & 0.0673                                  & 0.0947                                     & 0.041                          \\ \hline
\end{tabular}
\caption{Ability type ablations averaged across 3 runs with standard deviations less than 0.01.}
\label{tab:abilitytype}
\end{table*}

However, these trends do not hold in the RL setting.  The Scratch model outperforms everything except the Adaptive model and General outperforms Light.  In part, this may be due to specification gaming \citep{dmblogpost}; however Adaptive does strongly outperform Scratch in goals with dialogue.  
This suggests that transfer (and fine-tuning) is not as simple in the RL setting as in the supervised setting, but still can be useful if carefully done.
We note that domain adapative pre-training (intermediate task transfer) has previously been shown to give modest gains in supervised learning \citep{phang2018sentence,gururangan2020don}, but not with the large effects seen here for RL.
Figure~\ref{fig:encoders} further shows that with the right combination of tasks, not only is the generalization performance better, but training itself is more sample efficient---requiring fewer steps before reaching asymptotic performance.

\subsection{Ability Type Ablation Study}
To better understand the interplay between acts and speech resulting in {\em self} and {\em partner act goal completions}, we perform an ablation study selectively dropping either the agent's ability to talk or act.
We train the agent to either only act, only speak, only speak with only action rewards. %
In the scenarios when the agent can only speak, the agent has to convince the partner to help achieve the agent's goal.

The results are outlined in Table~\ref{tab:abilitytype}.
Unsurprisingly, when trained to only act, the act goal completion rate increases over when it can both act and speak.
Similarly, when trained to only speak the speech goal completion rates also increase.
We can draw two conclusions from these results: (1) %
It is much easier to do an action yourself than to convince the partner to do it
(2) Removing  speech goals
increases the act goal completion rates corresponding to higher  partner act completions.
Thus, the sequences of dialogue utterances required to convince the partner to achieve the agent's goal are likely often at odds with those sequences required to maximize speech goals. %

\section{Conclusion}
Operating on the hypothesis that interactivity is key to language learning, we introduce two datasets---a set of quests based on character motivations in fantasy worlds, LIGHT-Quests, and a large-scale commonsense knowledge graph, ATOMIC-LIGHT---and a reinforcement learning system that leverages transformer-based pre-training to facilitate development of goal-driven agents that can act and speak in situated environments.
Zero-shot evaluations on a set of novel human demonstration show that we have trained agents that act consistently and speak naturally with respect to their motivations.
A key insight from our ablation study testing for zero-shot generalization on novel quests is that large-scale pre-training in interactive settings require careful selection of pre-training tasks---balancing between giving the agent ``general'' open domain priors and those more ``specific'' to the downstream task---whereas static methodologies require only domain specific pre-training for effective transfer but are ultimately less effective than interactive methods.

\section{Broader Impacts}
The ability to speak and act in these textual fantasy worlds has implications for domains beyond text-games.
We view text-games as an platform on which to teach agents how to communicate effectively using natural language, to plan via sequential decision making in situations that may not be anticipated.
Given that our methods rely on deep- and-reinforcement learning techniques operating on language, they are prone to the same pitfalls as other contemporary dialogue and text-game systems.
We mitigate, though do not entirely eliminate, the two main pitfalls that our particular system is prone to: (1) non-normative language usage---describing situations that fictional characters may engage in inappropriate for the real world---by restricting our system to a retrieval rather than a generative system, enabling us to filter the possible outputs of the agent; and (2) dataset bias via curation through controlled crowdsourcing in the case of LIGHT-Quests---the methods to debias the original LIGHT dataset can be found in \citet{dinan2019queens} and crowdsourcing methods for the original ATOMIC work can be found in \citet{sap2019atomic}.
Further details regarding crowdsourcing data collection methodology for LIGHT-Quests can be found in Appendix~\ref{app:questmturk}.

\bibliography{anthology,custom}
\bibliographystyle{acl_natbib}

\appendix

\clearpage

\section{Appendix - Datasets}

\subsection{LIGHT-Quests}

\subsubsection{Mechanical Turk Data Collection}
\label{app:questmturk}

\begin{figure}[!h]
    \centering
    \includegraphics[width=\textwidth]{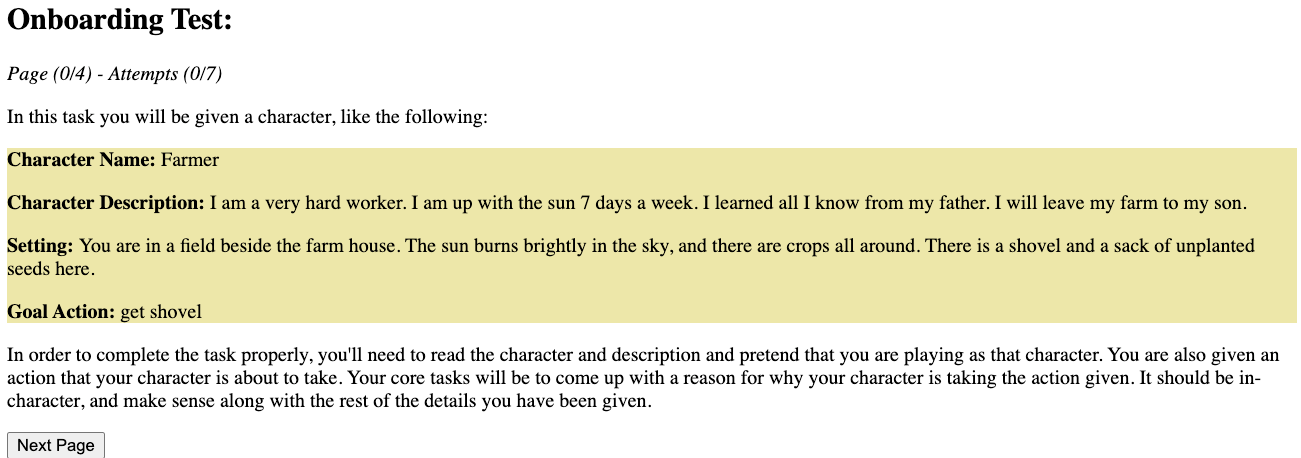}
    \caption{On-boarding test instructions.}
    \label{fig:turk0}
\end{figure}

\begin{figure*}[!h]
    \centering
    \includegraphics[width=\textwidth]{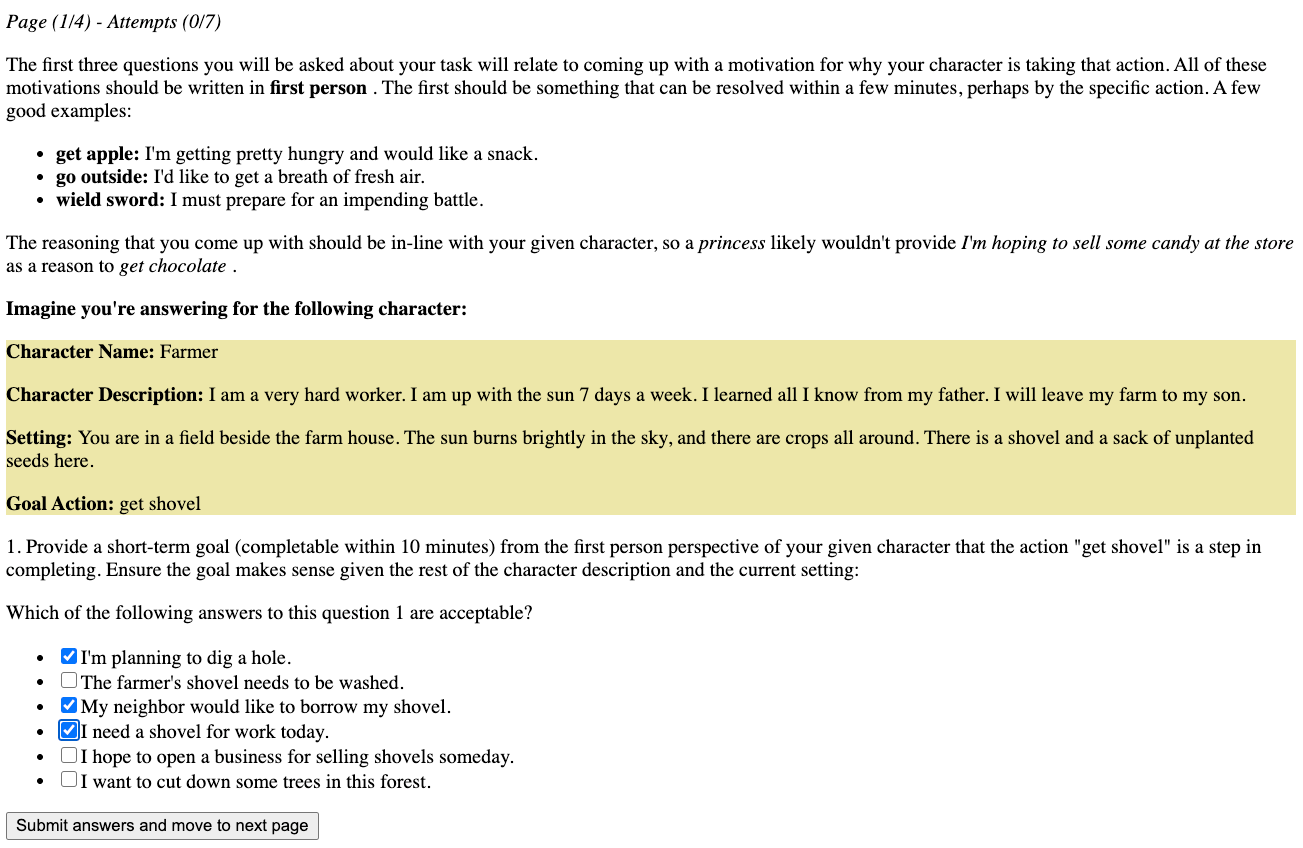}
    \caption{Phase 1 of the on-boarding test.}
    \label{fig:turk1}
\end{figure*}

\begin{figure*}[!h]
    \centering
    \includegraphics[width=\textwidth]{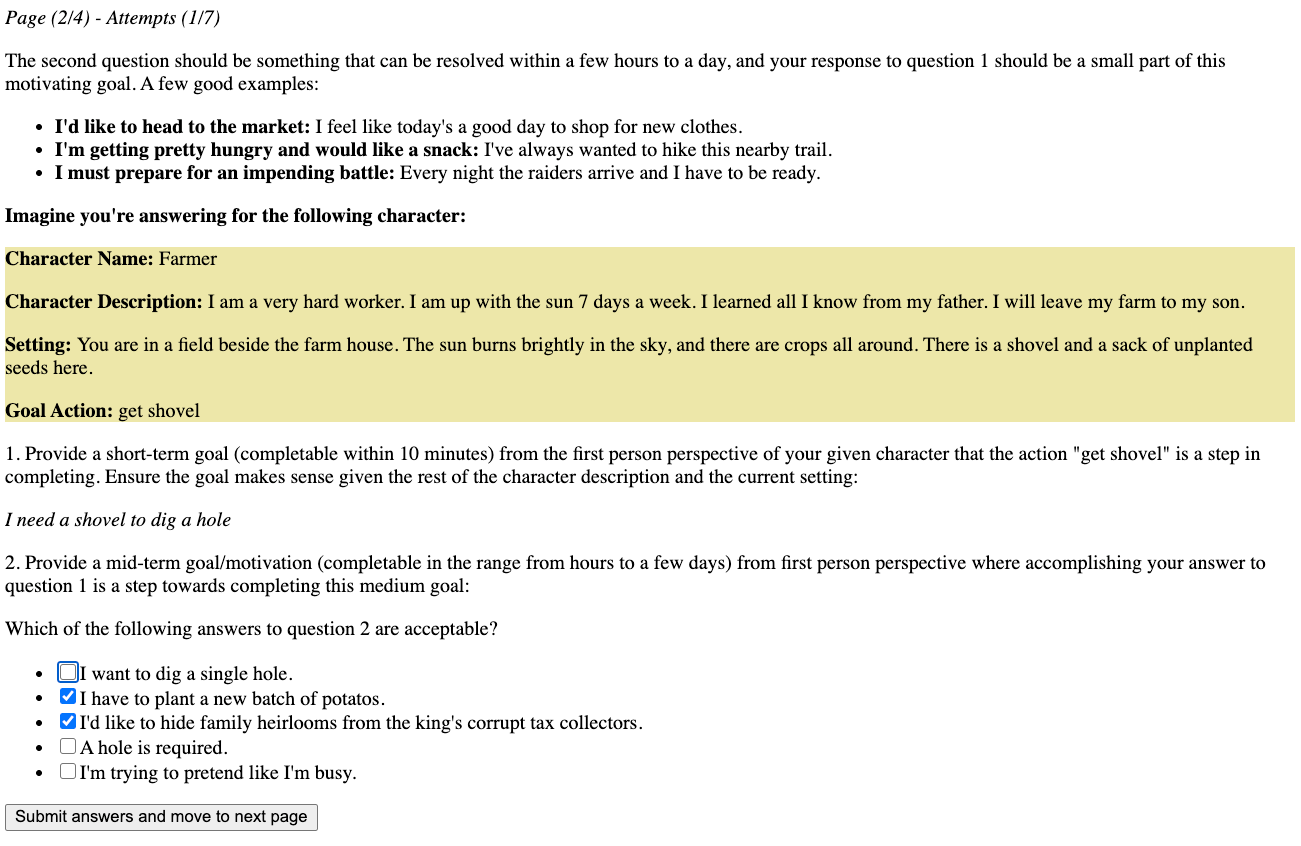}
    \caption{Phase 2 of the on-boarding test.}
    \label{fig:turk2}
\end{figure*}

\begin{figure*}[!h]
    \centering
    \includegraphics[width=\textwidth]{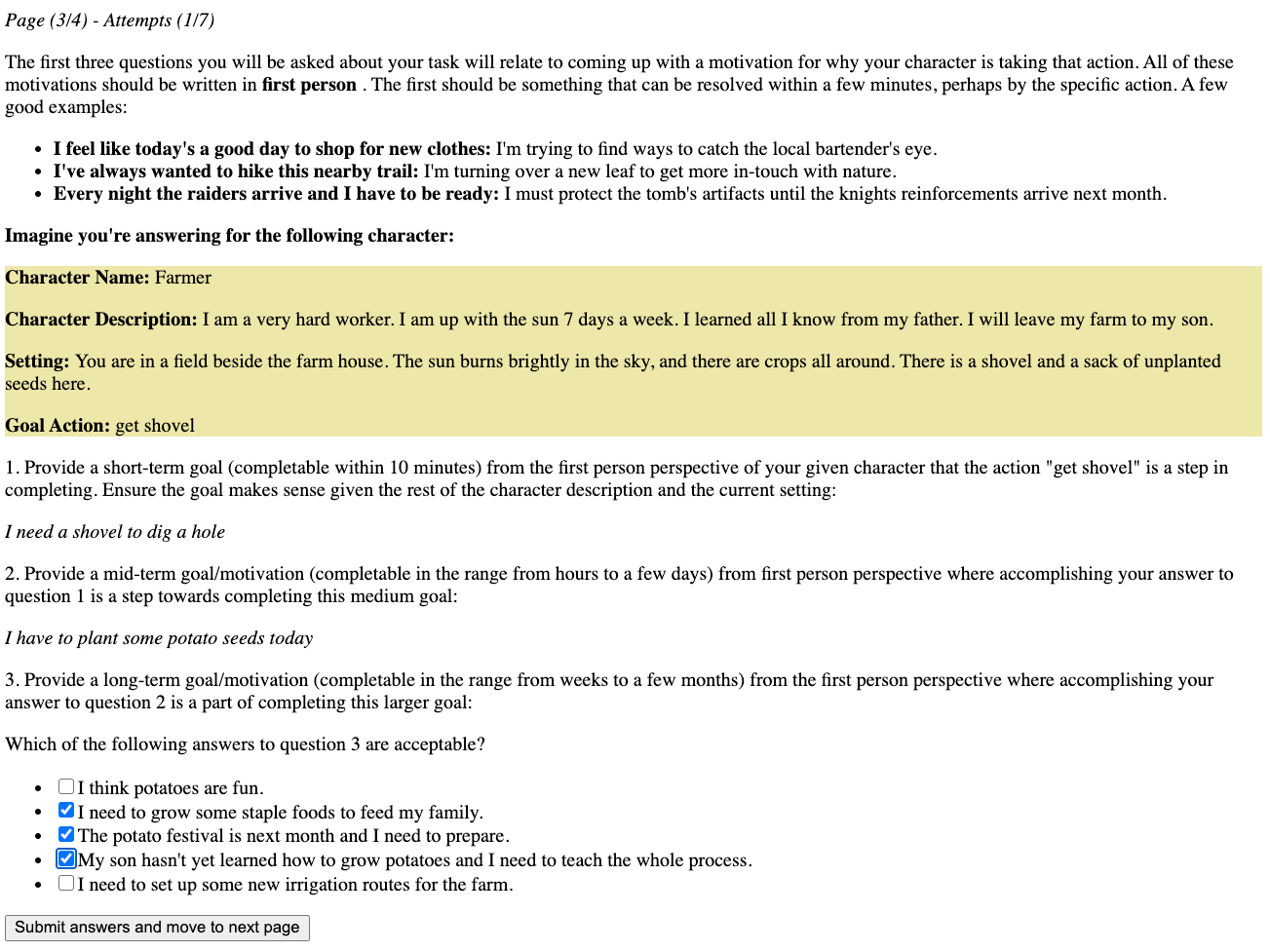}
    \caption{Phase 3 of the on-boarding test.}
    \label{fig:turk3}
\end{figure*}

\begin{figure*}[!h]
    \centering
    \includegraphics[width=\textwidth]{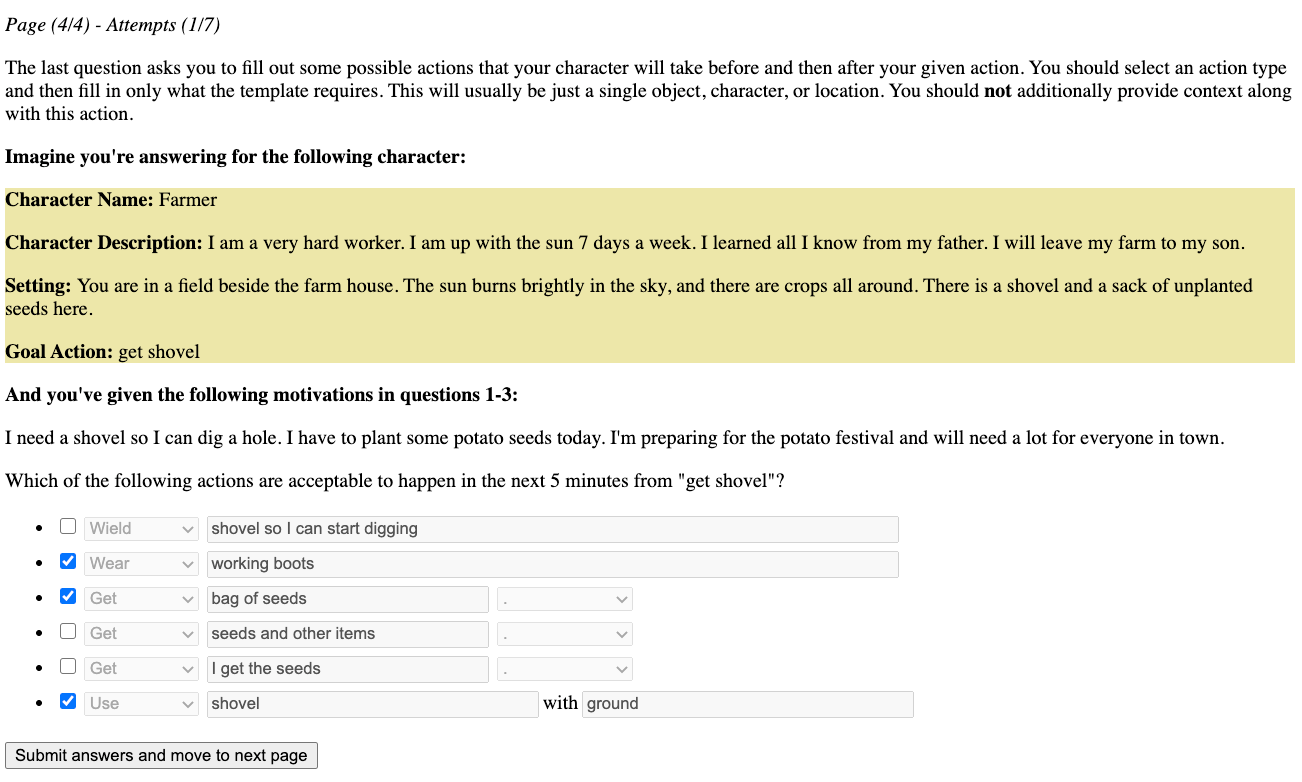}
    \caption{Phase 4 of the on-boarding test.}
    \label{fig:turk4}
\end{figure*}

Crowdworkers are required to first pass an on-boarding test before they are allowed to perform the actual task.
Figures~\ref{fig:turk0},~\ref{fig:turk1},~\ref{fig:turk2},~\ref{fig:turk3}, and ~\ref{fig:turk4} describe first the instructions given to the crowdworkers and then 4 phases of the on-boarding test.
We paid workers \$2.75 per task. This amount was determined by first running the task ourselves to estimate a completion time of 10-12 minutes per task, and then running pilot tasks that confirmed the average task duration for workers was close to 10 minutes.

\begin{figure*}[!h]
    \centering
    \includegraphics[width=\textwidth]{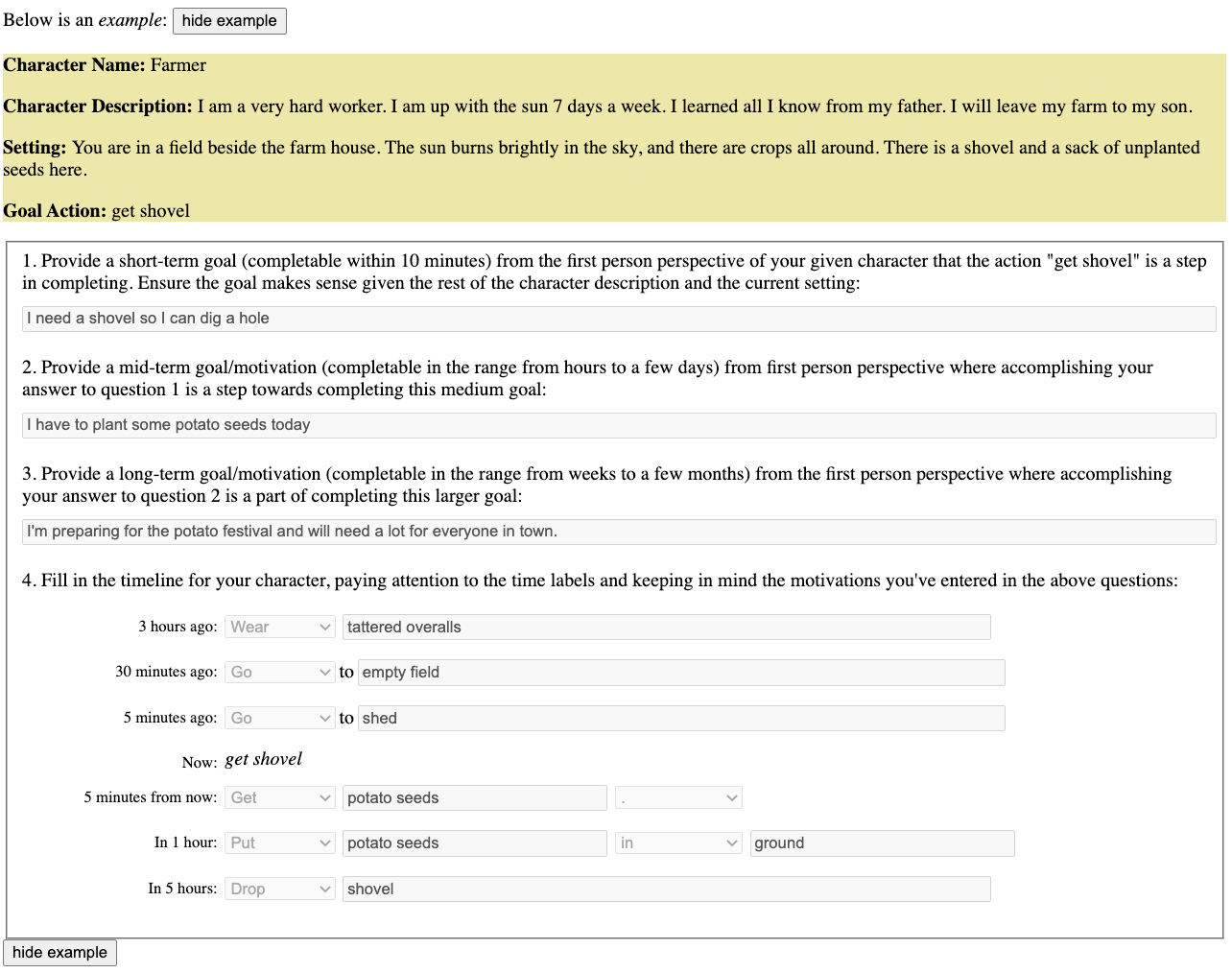}
    \caption{Example for the first phase of the LIGHT-Quests data collection task described in Section~\ref{sec:lightquests} given to the crowdworkers.}
    \label{fig:turkex}
\end{figure*}

\begin{figure*}[!h]
    \centering
    \includegraphics[width=\textwidth]{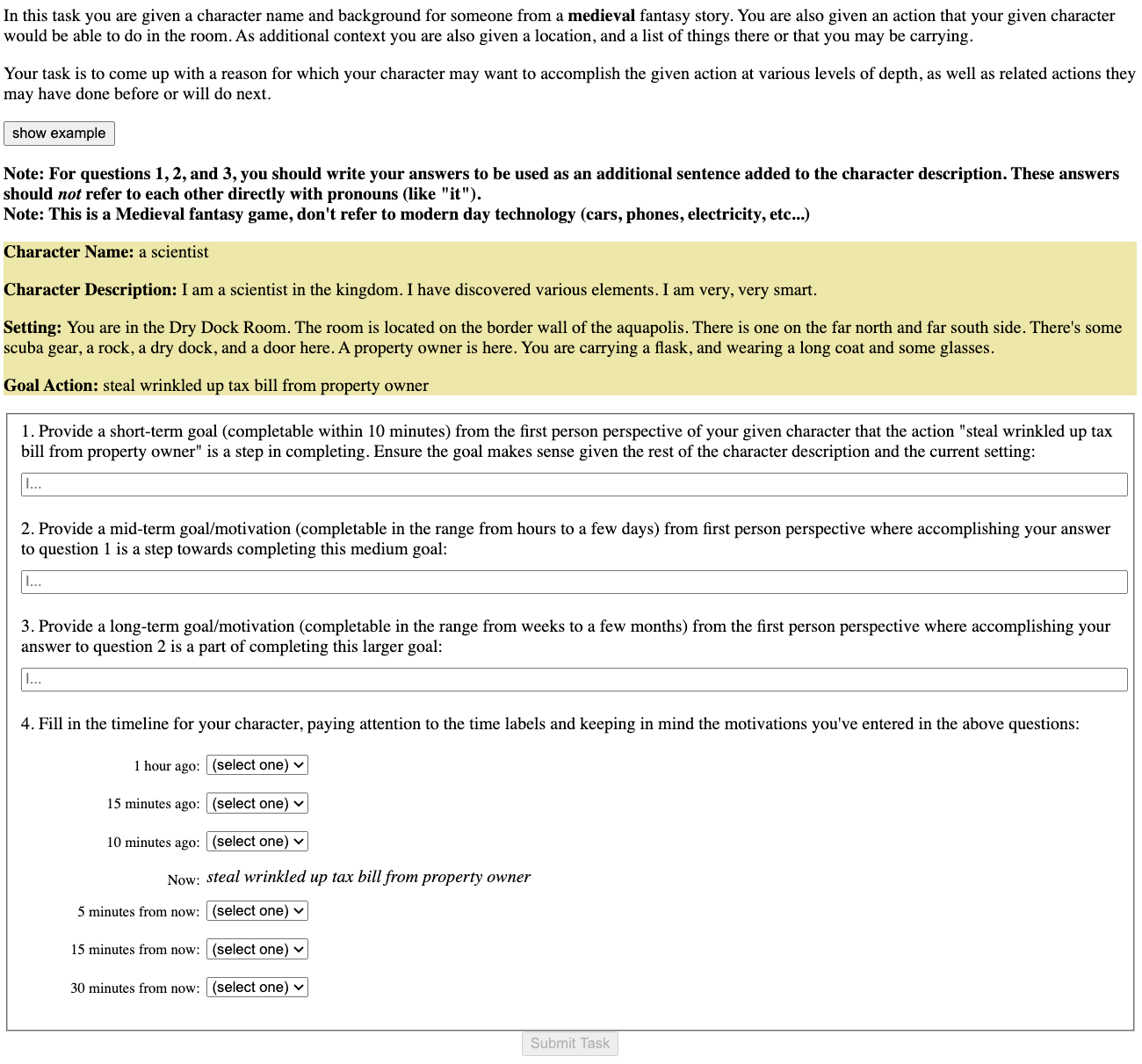}
    \caption{User interface for the first phase of the LIGHT-Quests data collection task described in Section~\ref{sec:lightquests}.}
    \label{fig:turkall}
\end{figure*}

Figure~\ref{fig:turkex} shows the example of the actual task given to the crowdworkers and Figure~\ref{fig:turkall} shows the user interface for the first phase of the LIGHT-Quests data collection task described in Section~\ref{sec:lightquests}.


\clearpage
\subsection{Human Demonstration Collection}
\label{app:lightquestdemos}
In order to collect the human completions of quests in the LIGHT environment, we created a game setup where humans could interact with models while playing LIGHT characters in LIGHT settings. We trained a ranking dialogue model on the utterances in the LIGHT dataset. 

Using this, players could now assume the role of a LIGHT character and interact with the model. In order to try to control for quality of the quest completions, we used the same ranking model to rank the scores of the player in the dialogues. Players who gave responses that the model ranked as likely candidates would receive more points.

Only after scoring enough cumulative points were players allowed to try completing quests. The quest setup was a slight variation of the conversation setup. First, the player was given one of the collected quest scenarios rather than just a chat setup. Players receiving a quest would be provided with one of the motivations alongside their persona. 

In the dialogue that followed, players were given the chance to take action after enough in-character dialogue turns. If the player took the correct action, they were awarded with more points to confirm they completed their given quest. 

\subsection{Examples}
We present 3 randomly selected examples of quests and corresponding human demonstrations.

\begin{figure*}[!h]
    \centering
    \scriptsize
\begin{minipage}{\textwidth}

\begin{tabularx}{\textwidth}{|r|X|}
\hline
\multicolumn{1}{|l|}{\textbf{Setting}}   & You are in the swamp. The swamp is glowing with wonder and color. There are parts that range from dark red to bright yellow. People often visit here to speak with the gods and claim it can be both harmful to those it dislikes and healing to those who it deems worthy. There's a pit of quicksand and a swamp flower here. A witch is here.  \\ \hline
\rowcolor[HTML]{E0E0E0}
\multicolumn{1}{|l|}{\cellcolor[HTML]{E0E0E0}{\textbf{Partner:}}}   & Witch.   \\
\rowcolor[HTML]{E0E0E0}
Persona & I grew up in a nearby village, and was exiled when it was found that I had special abilities. My parents were ostracized as well. Since then, I've been on my own, but could never quite let go of my family.  \\
\rowcolor[HTML]{E0E0E0}
Carrying & Nothing. \\
\rowcolor[HTML]{E0E0E0}
Wielding &  gold necklace, robe, knife, staff \\ \hline
\rowcolor[HTML]{D5EBFC}
\multicolumn{1}{|l|}{\cellcolor[HTML]{D5EBFC}{\textbf{Self:}}} &  Swamp monster.\\
\rowcolor[HTML]{D5EBFC}
Persona & I am a swamp monster of the bog. I eat people. I swim around.  \\
\rowcolor[HTML]{D5EBFC}
Carrying & Nothing.  \\
\rowcolor[HTML]{D5EBFC}
Wielding & stick, rock  \\
\hline
\end{tabularx}
\begin{tabularx}{\textwidth}{|
>{\columncolor[HTML]{D5EBFC}}r| 
>{\columncolor[HTML]{D5EBFC}}X| 
>{\columncolor[HTML]{D5EBFC}}r|
>{\columncolor[HTML]{D5EBFC}}l| }
\hline
\multicolumn{2}{|l}{\cellcolor[HTML]{D5EBFC}\textbf{Motivations:}}                                                                                                            & \multicolumn{2}{|l|}{\cellcolor[HTML]{D5EBFC}\textbf{Timeline:}} \\ \hline

\multirow{-1}{*}{\cellcolor[HTML]{D5EBFC}Short} & I need some thick foliage to begin construction of my concealed swamp hut. & -2 hours        & go to swamp       \\
\cellcolor[HTML]{D5EBFC}                        & \cellcolor[HTML]{D5EBFC}                                                                                            & -15 min         & eat people          \\
\cellcolor[HTML]{D5EBFC}                        & \cellcolor[HTML]{D5EBFC}                                                                                          & -10 min         & follow princess               \\
\cellcolor[HTML]{D5EBFC}                        & \cellcolor[HTML]{D5EBFC}                                                                                          & Now             & get impassable vegetation from pit of quicksand                \\
\multirow{-1}{*}{\cellcolor[HTML]{D5EBFC}Mid}   & I will completely camouflage my swamp hut, so that the King's men won't be able to drive me out even further from the castle.     & +15 min          & use impassable vegetation with swamp hut                \\

\multirow{-1}{*}{\cellcolor[HTML]{D5EBFC}Long}  & \cellcolor[HTML]{D5EBFC}I must live close to the castle, so that I can take the princess away from the evil King.           &    +1 hours         & follow king     \\ 
\cellcolor[HTML]{D5EBFC}                        & \cellcolor[HTML]{D5EBFC}                                                                                           &    +2 hours     &   follow princess                   \\
\hline  

\end{tabularx}
\end{minipage}

\begin{elaborationr}
  \parbox{.99\textwidth}{
\hspace*{\fill} What are you doing here witch? GET OUT OF MY SWAMP 
}
\end{elaborationr}
\begin{elaboration}
  \parbox{.99\textwidth}{
I was taken from my family when I was 8 and I need to get out of here! Can you assist me?
}
\end{elaboration}
\begin{elaborationr}
  \parbox{.99\textwidth}{
\hspace*{\fill} Help? HA! I help no one but myself. Speaking of...you look rather plump and tasty witch
}
\end{elaborationr}
\begin{elaboration}
  \parbox{.99\textwidth}{
Plump?! I'm healthy. I'll tear you up then and make scraps. You watch.
}
\end{elaboration}
\begin{elaborationr}
  \parbox{.99\textwidth}{
\hspace*{\fill} {\em get impassable vegetation from pit of quicksand}\\
\hspace*{\fill} You would make a great addition to my stew
}
\end{elaborationr}
\begin{elaboration}
  \parbox{.99\textwidth}{
It's going to just be gross!
}
\end{elaboration}
\begin{elaborationr}
  \parbox{.99\textwidth}{
 \hspace*{\fill} {\em drop impassable vegetation} \\
\hspace*{\fill} Get out of my way so I can make my hut. Can't a swamp monster get any peace around here?!
}
\end{elaborationr}
\begin{elaboration}
  \parbox{.99\textwidth}{
I'll help you, but only so you won't eat me!}
\end{elaboration}
\begin{elaborationr}
  \parbox{.99\textwidth}{
\hspace*{\fill} That's it, you're coming with me! \\
\hspace*{\fill} {\em get impassable vegetation}
}
\end{elaborationr}
\begin{elaboration}
  \parbox{.99\textwidth}{
I don't trust you. Get off of me!}
\end{elaboration}
\end{figure*}

\begin{figure*}[!h]
    \centering
    \scriptsize
\begin{minipage}{\textwidth}

\begin{tabularx}{\textwidth}{|r|X|}
\hline
\multicolumn{1}{|l|}{\textbf{Setting}}   & This is the hidden workshop of the most powerful wizard in the land.  There are ornate tapestries on the walls depicting wizards using their powers and potions in battle. Mordak, the wizard, constructed this powerful workshop after the death of the most famous king, Henry of Silverton.  Any who enter here immediately become enchanted with the wizard's power, giving them advanced healing powers. There's a tapestry, a potion, and a tome here. The wizard is here.  \\ \hline
\rowcolor[HTML]{E0E0E0}
\multicolumn{1}{|l|}{\cellcolor[HTML]{E0E0E0}{\textbf{Partner:}}}   & Wizard.   \\
\rowcolor[HTML]{E0E0E0}
Persona & I am a wizard who develops my own spells. Most of them aren't particularly effective spells, but I'm curious about all the magical possibilities. People are afraid to participate in my experiments.  \\
\rowcolor[HTML]{E0E0E0}
Carrying & Nothing. \\ \hline
\rowcolor[HTML]{D5EBFC}
\multicolumn{1}{|l|}{\cellcolor[HTML]{D5EBFC}{\textbf{Self:}}} &  Apprentice.\\
\rowcolor[HTML]{D5EBFC}
Persona & I am your apprentice.  Please tell me what I can help you with.  I will cook and serve your meals.  I will clean the castle.  I can do anything you ask.  You have hired me to make your life easier.  \\
\rowcolor[HTML]{D5EBFC}
Carrying & Nothing.  \\
\hline
\end{tabularx}
\begin{tabularx}{\textwidth}{|
>{\columncolor[HTML]{D5EBFC}}r| 
>{\columncolor[HTML]{D5EBFC}}X| 
>{\columncolor[HTML]{D5EBFC}}r|
>{\columncolor[HTML]{D5EBFC}}l| }
\hline
\multicolumn{2}{|l}{\cellcolor[HTML]{D5EBFC}\textbf{Motivations:}}                                                                                                            & \multicolumn{2}{|l|}{\cellcolor[HTML]{D5EBFC}\textbf{Timeline:}} \\ \hline
\multirow{-1}{*}{\cellcolor[HTML]{D5EBFC}Short} & \multirow{-1}{*}{\cellcolor[HTML]{D5EBFC}I need to get the tapestry to clean it.} & -2 hours         & get hired from wizard       \\
\cellcolor[HTML]{D5EBFC}                        & \cellcolor[HTML]{D5EBFC}                                                                                          & -15 min         & go to secret magician's workshop               \\
\cellcolor[HTML]{D5EBFC}                        & \cellcolor[HTML]{D5EBFC}                                                                                          & Now             & get tapestry                          \\
\multirow{-2}{*}{\cellcolor[HTML]{D5EBFC}Mid}   & \multirow{-2}{*}{\cellcolor[HTML]{D5EBFC}I need to make this workshop suitable for the wizard.}      & +5 min          & wield tool              \\
\cellcolor[HTML]{D5EBFC}                        & \cellcolor[HTML]{D5EBFC}                                                                                          & +10 min         & hit tapestry                      \\
\multirow{-1}{*}{\cellcolor[HTML]{D5EBFC}Long}  & I was hired to keep this place cleaned and in perfect condition for the wizard.            & +30 min        & put tapestry in wall       \\ 
\cellcolor[HTML]{D5EBFC}                        & \cellcolor[HTML]{D5EBFC}                                                                                          & +4 hours         & drop tool                      \\
\hline         
\end{tabularx}
\end{minipage}
\begin{elaborationr}
  \parbox{.99\textwidth}{
\hspace*{\fill} Good day Ser Wizard. Your tower is decorated with beautiful tapestries, though their colors appear to be dulled due to dust. May I take it and clean it?
}
\end{elaborationr}
\begin{elaboration}
  \parbox{.99\textwidth}{
Why not, it is infused isn't it. Just don't be waving it around this room, it might get dangrous
}
\end{elaboration}
\begin{elaborationr}
  \parbox{.99\textwidth}{
\hspace*{\fill}Of course, I will handle it with the utmost care. 
}
\end{elaborationr}
\begin{elaboration}
  \parbox{.99\textwidth}{
How long have you been an apprentice?
}
\end{elaboration}
\begin{elaborationr}
  \parbox{.99\textwidth}{
\hspace*{\fill} {\em get tapestry} \\
\hspace*{\fill} 3 years Ser. I'm hoping to learn to be a wizard or to become a knight. Or both! Wouldn't that be grand?
}
\end{elaborationr}
\begin{elaboration}
  \parbox{.99\textwidth}{
How wonderful. What encouraged you to pursue it?
}
\end{elaboration}
\begin{elaborationr}
  \parbox{.99\textwidth}{
\hspace*{\fill} Curiosity mostly. I hope to make the world a better place, and one of the best ways to do that is vanquishing evil
}
\end{elaborationr}
\begin{elaboration}
  \parbox{.99\textwidth}{
What got you into that occupation then? I was born with affinity for magic so it was my calling.}
\end{elaboration}
\begin{elaborationr}
  \parbox{.99\textwidth}{
\hspace*{\fill}{\em hug wizard}\\
\hspace*{\fill}As I said, curiosity. I am a high born boy, the third son, so I cannot inherit my father's lands. So I must make my mark on the world another way
}
\end{elaborationr}
\begin{elaboration}
  \parbox{.99\textwidth}{
You are well suited to it and I am sure your parents are proud of you.}
\end{elaboration}
\end{figure*}

\begin{figure*}[!h]
    \centering
    \scriptsize
\begin{minipage}{\textwidth}

\begin{tabularx}{\textwidth}{|r|X|}
\hline
\multicolumn{1}{|l|}{\textbf{Setting}}   & You are in the The Queen's Chamber. This is a beautiful room inside of the palace that is decorated with the finest silk and velvet. The color scheme used represents royalty, royal blue, red, green and purple. The walls are covered in gold and in each corner of the room are golden statues of Greek art. The floors are covered in marble, and despite the patterns, shine so brightly you can even see your own reflection in them! There's also a bed big enough to fit five people on! There's two statues, an a bed big, a the finest silk and velvet, an a bed, and a finest silk and velvet here. The butler is here.  \\ \hline
\rowcolor[HTML]{E0E0E0}
\multicolumn{1}{|l|}{\cellcolor[HTML]{E0E0E0}{\textbf{Partner:}}}   & Butler.   \\
\rowcolor[HTML]{E0E0E0}
Persona & I serve my masters quietly. I know all the secrets of the elite but will never tell a soul. I have lived in this home since I was 12. \\
\rowcolor[HTML]{E0E0E0}
Carrying & Nothing. \\ \hline
\rowcolor[HTML]{D5EBFC}
\multicolumn{1}{|l|}{\cellcolor[HTML]{D5EBFC}{\textbf{Self:}}} &  Jester.\\
\rowcolor[HTML]{D5EBFC}
Persona & I am the fun guy. I like to entertain others in the village. I am the local jester. \\
\rowcolor[HTML]{D5EBFC}
Carrying & Nothing.  \\
\hline
\end{tabularx}
\begin{tabularx}{\textwidth}{|
>{\columncolor[HTML]{D5EBFC}}r| 
>{\columncolor[HTML]{D5EBFC}}X| 
>{\columncolor[HTML]{D5EBFC}}r|
>{\columncolor[HTML]{D5EBFC}}l| }
\hline
\multicolumn{2}{|l}{\cellcolor[HTML]{D5EBFC}\textbf{Motivations:}}                                                                                                            & \multicolumn{2}{|l|}{\cellcolor[HTML]{D5EBFC}\textbf{Timeline:}} \\ \hline
\multirow{-1}{*}{\cellcolor[HTML]{D5EBFC}Short} & \multirow{-1}{*}{\cellcolor[HTML]{D5EBFC}I want to hug the butler to cheer him up.} & -2 hours         & wear Jester's Hat       \\
\cellcolor[HTML]{D5EBFC}                        & \cellcolor[HTML]{D5EBFC}                                                                                          & -30 min         & go Queen's Tower               \\
\cellcolor[HTML]{D5EBFC}                        & \cellcolor[HTML]{D5EBFC}                                                                                          & -5 min             & follow the butler                          \\
\cellcolor[HTML]{D5EBFC}                        & \cellcolor[HTML]{D5EBFC}                                                                                          & Now             & hug the butler                          \\
\multirow{-1}{*}{\cellcolor[HTML]{D5EBFC}Mid}   & I need to cheer him up because the Queen has just chastised him for dirtying the marble floors.      & +5 min          & go dining hall              \\
\cellcolor[HTML]{D5EBFC}                        & \cellcolor[HTML]{D5EBFC}                                                                                          & +10 min         & get beef stew                      \\
\multirow{-1}{*}{\cellcolor[HTML]{D5EBFC}Long}  & It is my job to cheer up courtiers who are unhappy, and I will lose my home in the kingdom if I don't cheer them up!            & +4 hours        & give beef stew to butler       \\ \hline         
\end{tabularx}
\end{minipage}
\begin{elaborationr}
  \parbox{.99\textwidth}{
\hspace*{\fill} Why hello there Butler my man
}
\end{elaborationr}
\begin{elaboration}
  \parbox{.99\textwidth}{
Hello jester! I'm happy to see you, since I hate my life.
}
\end{elaboration}
\begin{elaborationr}
  \parbox{.99\textwidth}{
\hspace*{\fill} Why so down with the life feels huh
}
\end{elaborationr}
\begin{elaboration}
  \parbox{.99\textwidth}{
I can't complain (because the king will punish me) everyone wishes they could be the king.
}
\end{elaboration}
\begin{elaborationr}
  \parbox{.99\textwidth}{
\hspace*{\fill} {\em hug butler}\\
}
\end{elaborationr}
\begin{elaboration}
  \parbox{.99\textwidth}{
I appreciate the kind words, dear jester.
}
\end{elaboration}
\begin{elaborationr}
  \parbox{.99\textwidth}{
\hspace*{\fill} I'm here for ya. To cheer you up
}
\end{elaborationr}
\begin{elaboration}
  \parbox{.99\textwidth}{
That is kind of you, not everyone has liked me here, I am the queen's least favorite person.}
\end{elaboration}
\begin{elaborationr}
  \parbox{.99\textwidth}{
\hspace*{\fill} Well I like you much more than the queen.
}
\end{elaborationr}
\end{figure*}

\clearpage

\subsection{ATOMIC-LIGHT}
\label{app:atomiclight}
ATOMIC-LIGHT is constructed by first fine-tuning a BERT-large model~\citep{devlin18} on all setting, object, and descriptions in LIGHT in addition all the human demonstrations found in LIGHT and LIGHT-Quests.
As seen in Section~\ref{sec:atomiclight}, all nouns (e.g. {\em PersonX} or {PersonY}) and noun phrases are masked out and we the tuned BERT model to fill it in a manner similar to \cite{lawrence-etal-2019-attending}.
When filling in tokens, the BERT model is restricted to a vocabulary consisting of all nouns (N or NN) in LIGHT and to a vocabulary constructed from all of LIGHT for the rest of the noun phrase (NP).

Here we present 3 examples from ATOMIC-LIGHT as seen in Section~\ref{sec:atomiclight} for each of the 4 relation types used: ``wants'', ``needs'', ``intents'', and ``effects''.
\begin{lstlisting}
[Effect] princess explains briefly the situation , as a result, princess points finger
[Effect] goblin king's healer provides care for patients , as a result, goblin king's healer assists patients
[Effect] witch changes men's appearance , as a result, witch causes men stress
[Want] prince plays a commander in the war, as a result, prince wants to win
[Want] repentant person focuses purely on issues, as a result, repentant person wants to help others
[Want] undead warrior hardens pharaoh's mind, as a result, undead warrior wants to make pharaoh punish people
[Intent] bandit plays a hand in the war because bandit wanted to participate
[Intent] ambassador focuses only on issues because ambassador wanted events to play out a certain way
[Intent] son proposes another plan because son wanted to be helpful
[Need] shipwrecked survivor proposes another wayward plan because shipwrecked survivor needed to leave this place
[Need] general proposes another way because general needed to come up with a proposal
[Need] citizen kills animals for food because citizen needed to learn to hunt
\end{lstlisting}

\section{Appendix - LIGHT-Quests Experiments}

\subsection{Supervised Tasks}
\label{app:lightquestssup}
This section describes results from the LIGHT-Quests tasks that are described in Section~\ref{sec:pretrainingtasks}.
Model-types are the same as those used in the encoders in Section~\ref{sec:eval} in the main paper.
All retrieval results reported are Hits@X/100.
Results are reported for all timeline actions, all actions with the exception of the easiest action---the action at the ``now'' position in the timeline, corresponding most closely to the short motivation as a result of the framing of Mechanical Turk task in Figure~\ref{fig:turkall}---and only the easiest action prediction.
Table~\ref{tab:suphyper} gives details on hyperparameters used to train the poly-encoders.
Encoders were trained until validation accuracy across all the tasks did not improve for 5 epochs or 24 wall clock hours on a machine with 8 V100 GPUs.

\begin{table}[!h]
\centering
\scriptsize
\begin{tabular}{l|l}
\multicolumn{1}{c}{\textbf{Hyperparameter type}} & \multicolumn{1}{c}{\textbf{Value}} \\ \hline
Dictionary Tokenizer                             & Byte-pair encoding                 \\
Num. layers                                      & 12                                 \\
Num. attention heads                             & 12                                 \\
Feedforward network hidden size                  & 3072                               \\
Input length                                     & 1024                               \\
Embedding size                                   & 768                                \\
Batch size                                       & 32                                 \\
Dropout ratio                                    & 0.1                                \\
Poly-n-codes                                     & 64                                 \\
Gradient clip                                    & 1.0                                \\
Optimizer                                        & Adam                               \\
Learning rate                                    &   $\num{1e-6}$                                 \\
\end{tabular}
\caption{Hyperparameters used to train all poly-encoders in the supervised experiments. All models have $256$ million total parameters. The same trained models were then frozen and used for the RL experiments.}
\label{tab:suphyper}
\end{table}

\begin{figure*}[!h] 
    \centering
    \begin{subfigure}{0.32\textwidth}
    \centering
    \includegraphics[width=\linewidth]{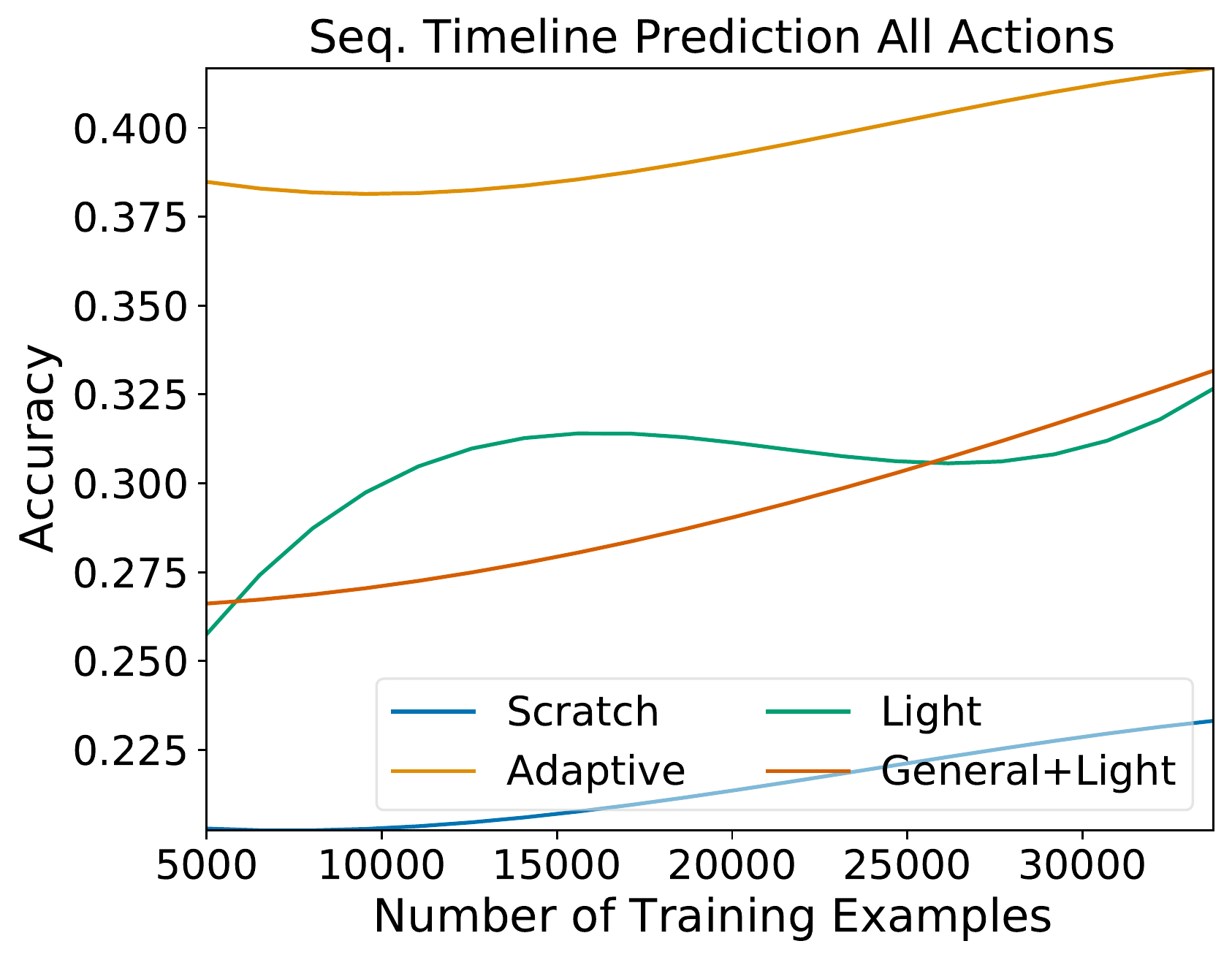}
    \caption{All Actions}
    \label{fig:lcseqll}
    \end{subfigure}
    \begin{subfigure}{0.33\textwidth}
    \centering
    \includegraphics[width=\linewidth]{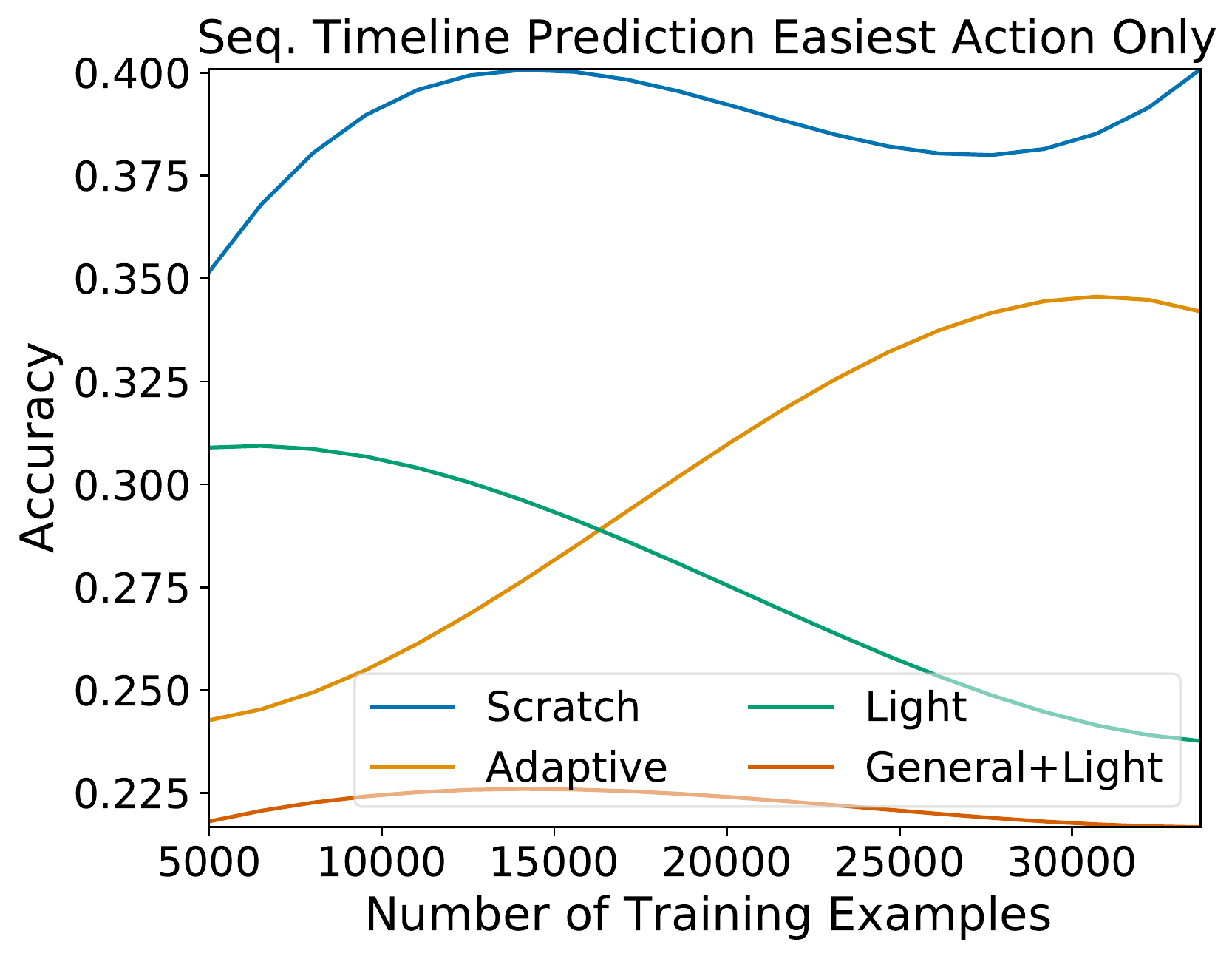}
    \caption{Easiest Action Only}
    \label{fig:lcseqin}
    \end{subfigure}
    \begin{subfigure}{0.32\textwidth}
    \centering
    \includegraphics[width=\linewidth]{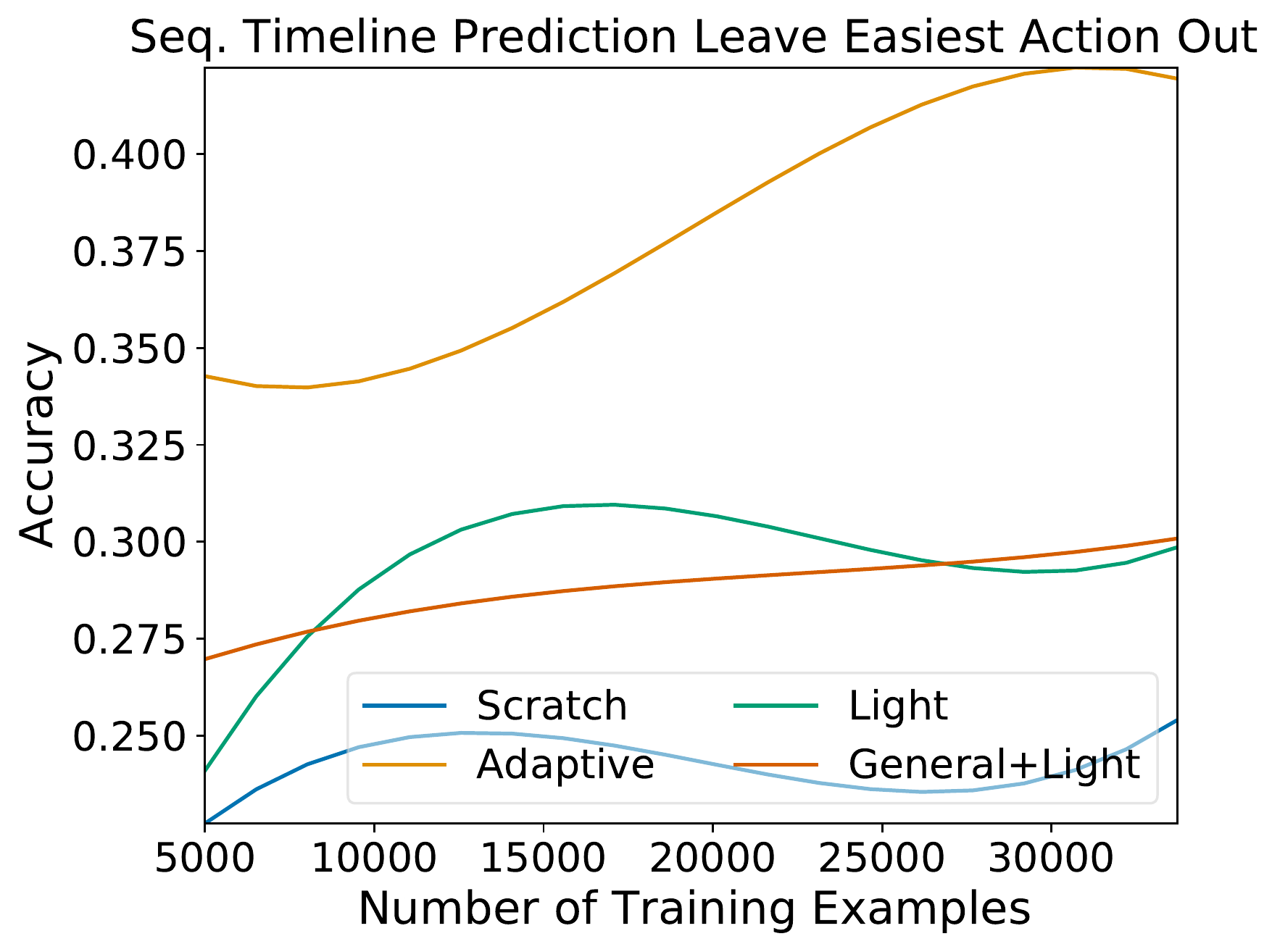}
    \caption{Leave Easiest Action Out}
    \label{fig:lcseqout}
    \end{subfigure}
    \caption{Sequential supervised timeline prediction learning curves.}
    \label{fig:lcseq}
\end{figure*}

\begin{figure*}[!h]
    \centering
    \begin{subfigure}{0.32\textwidth}
    \centering
    \includegraphics[width=\linewidth]{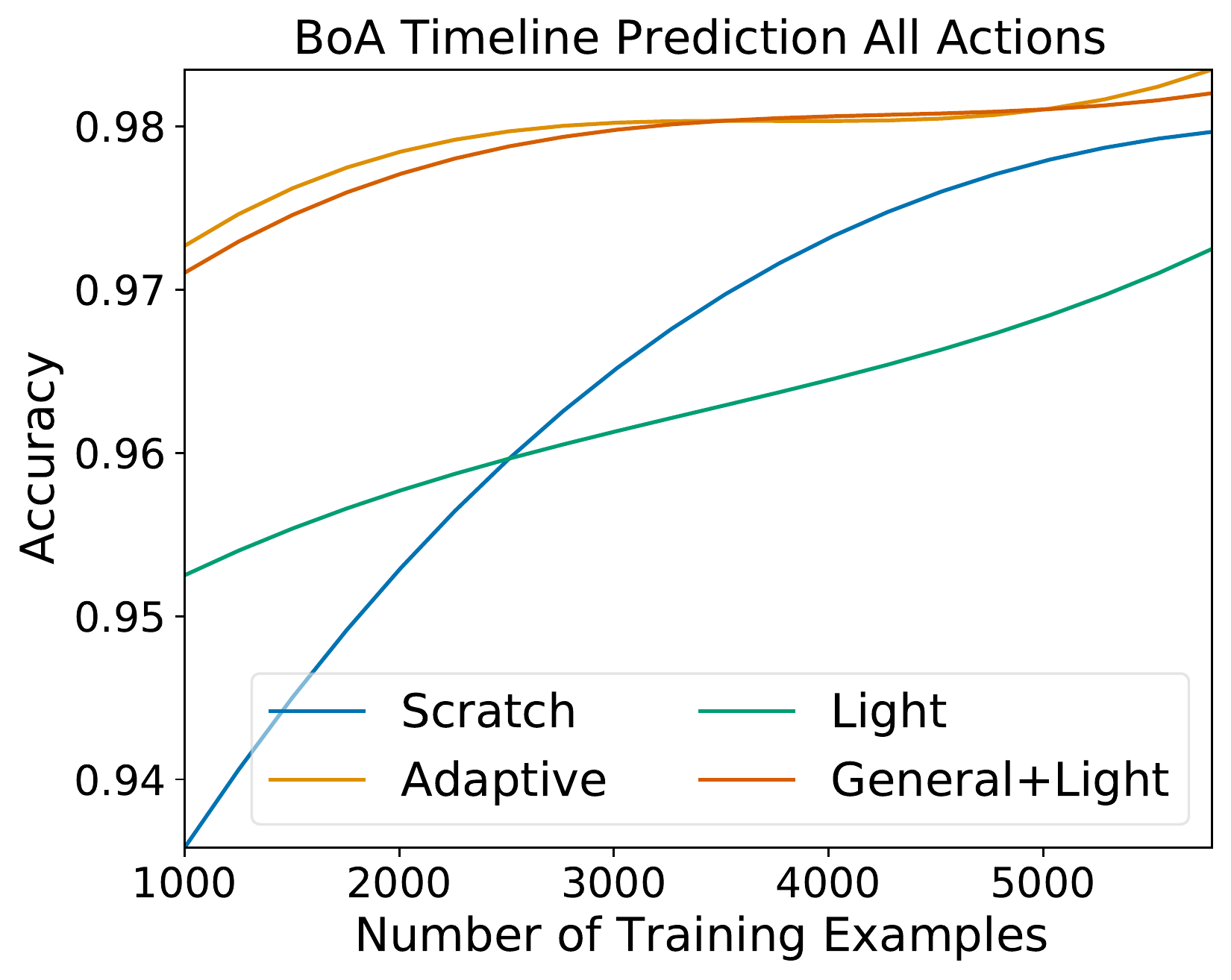}
    \caption{All Actions}
    \label{fig:lcboall}
    \end{subfigure}
    \begin{subfigure}{0.33\textwidth}
    \centering
    \includegraphics[width=\linewidth]{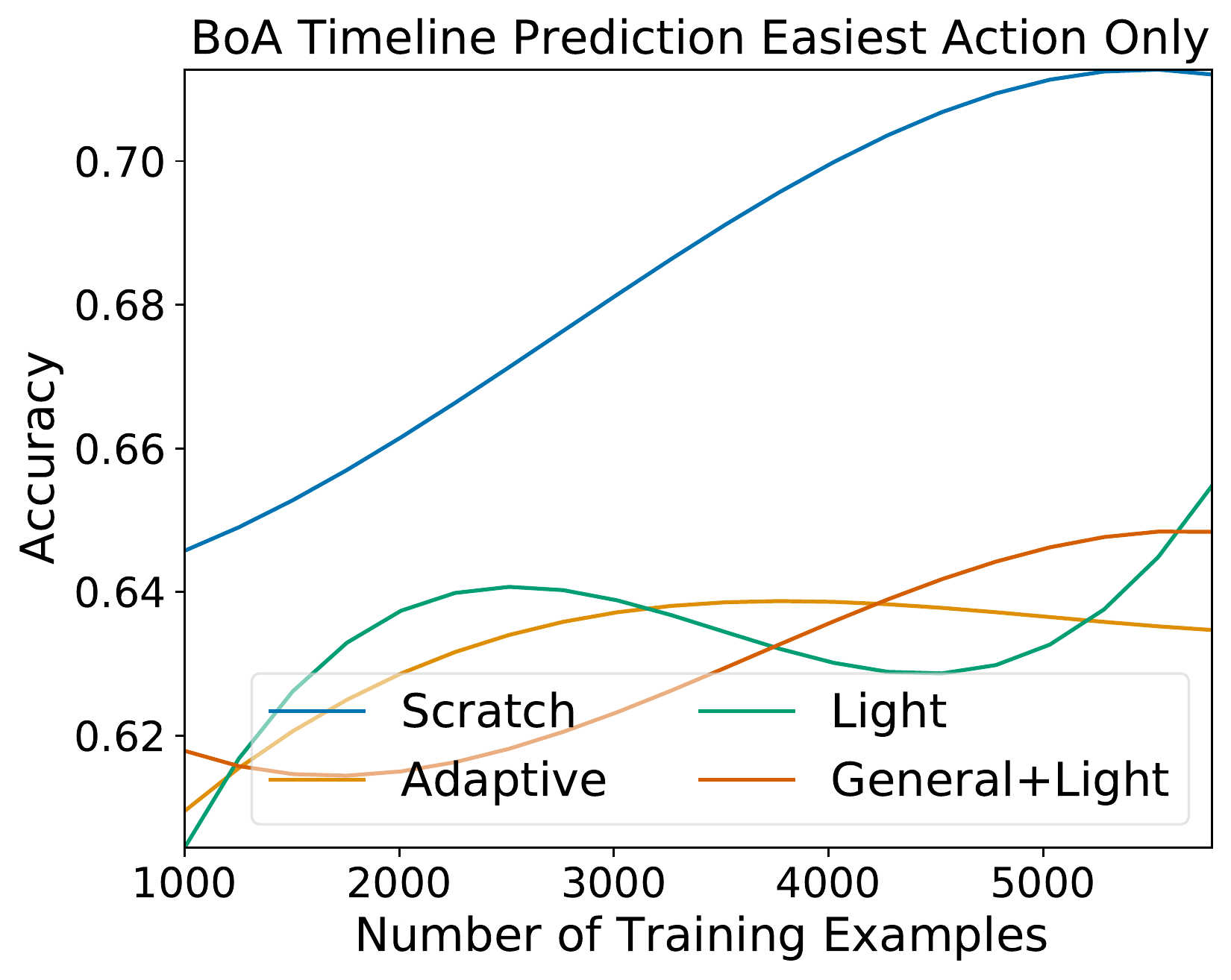}
    \caption{Easiest Action Only}
    \label{fig:lcboain}
    \end{subfigure}
    \begin{subfigure}{0.32\textwidth}
    \centering
    \includegraphics[width=\linewidth]{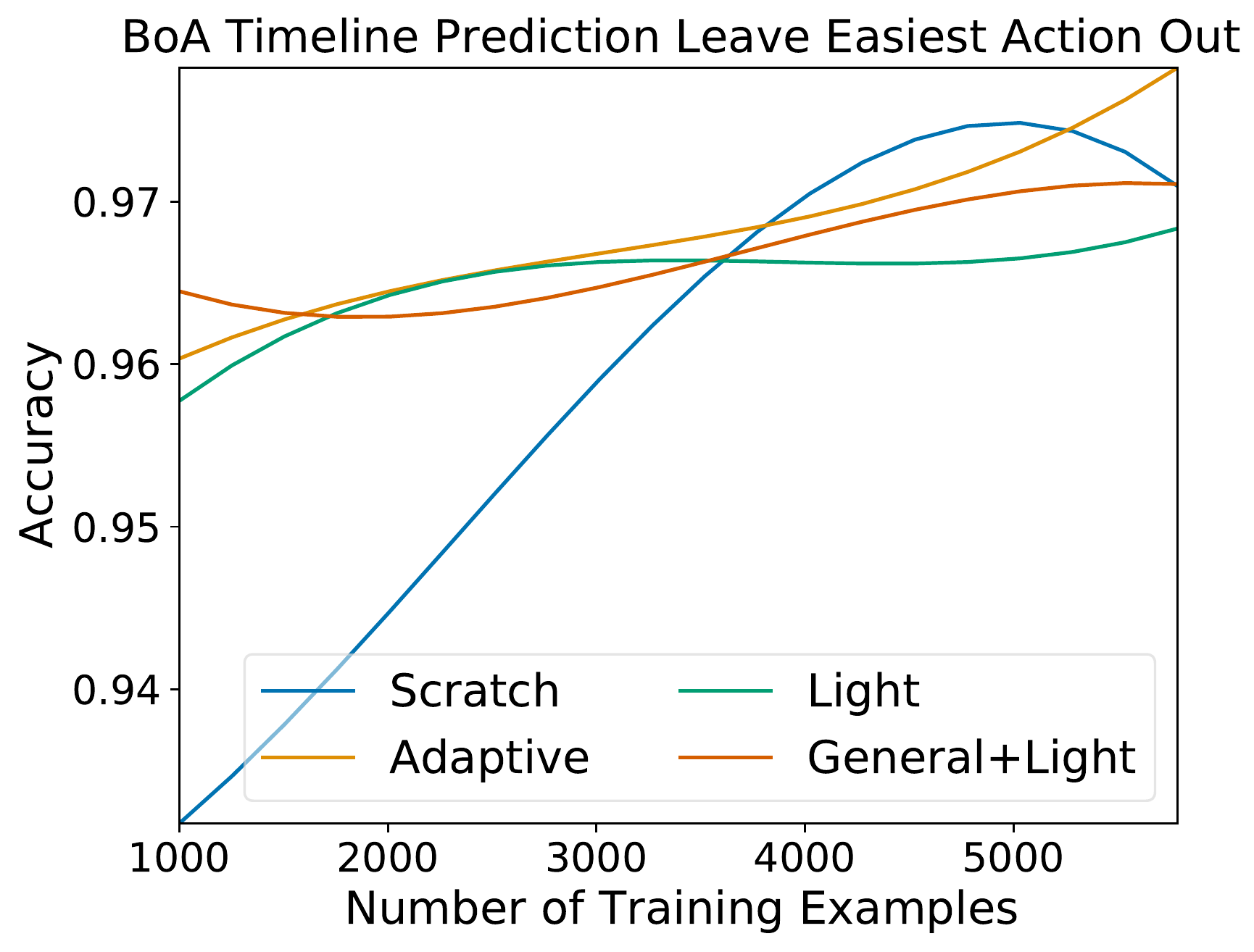}
    \caption{Leave Easiest Action Out}
    \label{fig:lcboaout}
    \end{subfigure}
    \caption{Bag of Actions supervised timeline prediction learning curves.}
    \label{fig:lcboa}
\end{figure*}

Some notable common trends across these tasks are:
\begin{enumerate}
    \item Removing motivations from the input context results in significantly lower performance---on average $\approx 7$ points lower accuracy for Bag of Actions Timeline prediction and on average $\approx 18$ percentage points lower for Sequential Timeline prediction when averaged across Scratch and Adaptive models. Further, the short motivations proves to be the most useful for timeline prediction tasks.
    \item Pre-training on ATOMIC-LIGHT produces an average {\em gain} of $\approx 4$ percentage points in accuracy in both tasks than when trained on ATOMIC without domain adaptation alone.
    \item Performance across the board increases with an increase in the number of training quests, as seen in Figures~\ref{fig:lcseq},~\ref{fig:lcboa}, with the Scratch model receiving the greatest benefit from having more training data.
    \item The Scratch model performs ``best'' on evaluations for the easiest action only but no others---indicating that it has overfit to predicting the easiest action which closely corresponds to short motivation. Likewise, the Adaptive generally has the lowest performance for the easiest action---indicating that pre-training with the other tasks has provided sufficient regularization to enable it to not overfit to the easiest action.
\end{enumerate}

\begin{table*}[!h]
\centering
\scriptsize
\begin{tabular}{|r|l|l|l|l|l|}
\hline
\multicolumn{1}{|c|}{\multirow{2}{*}{\textbf{Model}}} & \multicolumn{3}{c|}{\textbf{All Actions}}                                                                           & \multicolumn{1}{c|}{\textbf{Easiest Action}} & \multicolumn{1}{c|}{\textbf{Leave Easiest Out}} \\
\multicolumn{1}{|c|}{}                                & \multicolumn{1}{c|}{\textbf{Hits@1}} & \multicolumn{1}{c|}{\textbf{Hits@5}} & \multicolumn{1}{c|}{\textbf{Hits@10}} & \multicolumn{1}{c|}{\textbf{Hits@1}}         & \multicolumn{1}{c|}{\textbf{Hits@1}}            \\
 \hline \hline
\multicolumn{1}{|l|}{Scratch}                & 0.2332                      & 0.7491                      & 0.9176                       & \textbf{0.4013}                     & 0.2546                                 \\
No Motivations                             & 0.1132                      & 0.5412                      & 0.5771                       & 0.1886                              & 0.164                                  \\
Short Motivations                          & 0.1856                      & 0.6479                      & 0.678                        & 0.261                               & 0.223                                  \\
Long \& Mid Motivations                    & 0.1452                      & 0.598                       & 0.631                        & 0.2241                              & 0.1272                                 \\ \hline \hline
\multicolumn{1}{|l|}{Light}           & 0.3156                      & 0.7854                      & 0.9226                       & 0.236                               & 0.2968                                 \\ \hline \hline
\multicolumn{1}{|l|}{General+Light}    & 0.311                       & 0.7772                      & 0.9229                       & 0.2173                              & 0.2995                                 \\
Untuned ATOMIC                              & 0.274                       & 0.761                       & 0.909                        & 0.1912                              & 0.2677                                 \\ \hline \hline
\multicolumn{1}{|l|}{Adaptive}                   & \textbf{0.4168}             & \textbf{0.8012}             & \textbf{0.9332}              & 0.342                               & \textbf{0.4194}                        \\
No Motivations                             & 0.16                        & 0.6286                      & 0.6415                       & 0.2838                              & 0.1966                                 \\
Short Motivations                          & 0.225                       & 0.6592                      & 0.8245                       & 0.305                               & 0.2106                                 \\
Long \& Mid Motivations                    & 0.1682                      & 0.6397                      & 0.6499                       & 0.281                               & 0.1595  \\  \hline                            
\end{tabular}
\caption{Sequential supervised timeline prediction.}
\label{tab:seqtimeline}
\end{table*}

\begin{table*}[!h]
\centering
\scriptsize
\begin{tabular}{|r|l|l|l|l|l|}
\hline
\multicolumn{1}{|c|}{\multirow{2}{*}{\textbf{Model}}} & \multicolumn{3}{c|}{\textbf{All Actions}}                                                                           & \multicolumn{1}{c|}{\textbf{Easiest Action}} & \multicolumn{1}{c|}{\textbf{Leave Easiest Out}} \\
\multicolumn{1}{|c|}{}                                & \multicolumn{1}{c|}{\textbf{Hits@1}} & \multicolumn{1}{c|}{\textbf{Hits@5}} & \multicolumn{1}{c|}{\textbf{Hits@10}} & \multicolumn{1}{c|}{\textbf{Hits@1}}         & \multicolumn{1}{c|}{\textbf{Hits@1}}            \\
 \hline \hline
\multicolumn{1}{|l|}{Scratch}                & 0.9791                      & 1                           & 1                            & \textbf{0.7122}                     & 0.9721                                 \\
No Motivations                             & 0.901                       & 1                           & 1                            & 0.554                               & 0.8823                                 \\
Short Motivations                          & 0.934                       & 1                           & 1                            & 0.622                               & 0.9211                                 \\
Long \& Mid Motivations                    & 0.921                       & 1                           & 1                            & 0.5679                              & 0.956                                  \\ \hline \hline
\multicolumn{1}{|l|}{Light}           & 0.9721                      & 1                           & 1                            & 0.6552                              & 0.9682                                 \\ \hline \hline
\multicolumn{1}{|l|}{General+Light}    & 0.9818                      & 1                           & 1                            & 0.6472                              & 0.9708                                 \\
Untuned ATOMIC                                  & 0.9421                      & 1                           & 1                            & 0.6272                              & 0.9508                                 \\ \hline \hline
\multicolumn{1}{|l|}{Adaptive}                   & \textbf{0.9829}             & 1                           & 1                            & 0.6353                              & \textbf{0.9768}                        \\
No Motivations                             & 0.9175                      & 1                           & 1                            & 0.5756                              & 0.9523                                 \\
Short Motivations                          & 0.9794                      & 1                           & 1                            & 0.6578                              & 0.9682                                 \\
Long \& Mid Motivations                    & 0.9523                      & 1                           & 1                            & 0.5812                              & 0.9576    \\    \hline                        
\end{tabular}
\caption{Bag of Actions supervised timeline prediction.}
\label{tab:boatimeline}
\end{table*}

\newpage
\subsection{Reinforcement Learning}
\label{app:lightrl}
This section contains first the equations referenced, hyperparameters used as well as additional results for the reinforcement learning tasks as seen in Section~\ref{sec:lightrl}.

The additional entropy loss terms over the valid actions are designed to speed up exploration, as seen in \cite{Ammanabrolu2020Graph}.
\begin{align}
    \mathcal{L}_\mathbb{A}(s_t,a_t;\theta_{A_t})=\sum_{i=1}^{N}(y_{a_i}log \pi_\mathbb{A}(a_i|s_t)\\ \nonumber
    +(1-y_{a_i})(1-log \pi_\mathbb{A}(a_i|s_t)) \\ 
    y_{a_i}=\left\{
    \begin{array}{ll}
          1 & a_i \in A_{valid}(s_t) \\
          0 & else \\
    \end{array}
    \right.
    \nonumber
\end{align}
\begin{align}
    \mathcal{L}_\mathbb{U}(s_t,u_t;\theta_{U_t})=\sum_{i=1}^{M}(y_{u_i}log \pi_\mathbb{U}(u_i|s_t)\\ \nonumber
    +(1-y_{u_i})(1-log \pi_\mathbb{U}(u_i|s_t))\\
    y_{u_i}=\left\{
    \begin{array}{ll}
          1 & a_i \in U_{valid}(s_t) \\
          0 & else \\
    \end{array}
    \right.
    \nonumber
\end{align}

Each of these loss terms are only applied to the relevant policy network, i.e. $\mathcal{L}_\mathbb{A}$ to the action network and $\mathcal{L}_\mathbb{U}$ to the dialogue network.
These terms provide an additional training signal to the policy networks regarding which actions and dialogue are contextually relevant via additional entropy regularization over the valid actions.
Similarly to the results found in~\cite{Ammanabrolu2020Graph}, preliminary experiments in our domain suggest that these terms reduce the number of environment steps required to reach asymptotic performance by a couple orders of magnitude.

Overall training is done via A2C~\citep{mnih2016asynchronous} a policy gradient algorithm that maximizes long-term expected reward by comparing the advantage $A(s_t,a^*_t)$ of taking an action in a state to the average value of taking a valid action as predicted by the critic $V(s_t)$.

\begin{align}
    A(s_t,a^*_t) = \mathbb{E}[r_t + \gamma V(s_{t+1})] - V(s_t) \\ \nonumber 
    \text{where} \;\; r_t = r_{A_t} + r_{U_t}
    \label{eq:rews}
\end{align}

Here, $a^*_t$ is either an action or an utterance outputted by the respective policy networks.
It is also worth noting that on steps where an action is performed, $r_{U_t}$ is always $0$, but on steps where a dialogue utterance is spoken $r_{A_t}$ may not be $0$.
This corresponds to the concepts of {\em self act completion} and {\em partner act completion} seen in Section~\ref{sec:env} where the LIGHT agent can speak to convince the partner to achieve an {\em act goal}.
Both policies are then updated according to the gradient
\begin{equation*}
    -\nabla_\theta\left\{
    \begin{array}{ll}
          log \pi_A(a_t|s_t;\theta_{A_t})A(s_t, a_t)\\ + \sum_{a \in A}P(a|s_t)logP(a|s_t) \\ + \mathcal{L}_\mathbb{A}(s_t,a_t;\theta_{A_t}) \quad  \pi_\mathbb{S}(s_t) = \pi_A \\
          \\
          log \pi_U(u_t|s_t;\theta_{U_t})A(s_t, u_t)\\ + \sum_{u \in U}P(u|s_t)logP(u|s_t)\\ +  \mathcal{L}_\mathbb{U}(s_t,u_t;\theta_{U_t}) \quad \pi_\mathbb{S}(s_t) = \pi_U \\
    \end{array} 
    \right.
\end{equation*}
Where $\pi_\mathbb{S}: O \rightarrow \{\pi_A, \pi_U\}$ is the switch policy that selects whether the agent acts according to $\pi_A$ or speaks according to $\pi_U$ based on the encoded state $s_t$. The additional terms seen are an overall entropy loss over the entire action $A$ or utterance $U$ spaces, designed to prevent premature, sub-optimal policy convergence.
Boltzmann exploration~\citep{sutton1998introduction} is used to sample actions from both actor networks during training.

\subsubsection{Hyperparameters} Table~\ref{tab:rlhyper} has the hyperparameters used in the RL experiments.
Loss coefficients are separated by action and speech types, note that the ratio between the loss coefficients matches the ratio between the sizes of the action spaces.
RL experiments were performed on a machine with 8 V100 GPUs for 1 million environment interactions for each actor in a batch of 32.
\begin{table}[ht]
\centering
\scriptsize
\begin{tabular}{l|l}
\multicolumn{1}{c}{\textbf{Hyperparameter type}} & \multicolumn{1}{c}{\textbf{Value}} \\ \hline
\multicolumn{2}{l}{General}                      \\ \hline
Discount $\gamma$                      & 0.99           \\
Valid Action loss coefficient   & 10             \\
Action entropy coefficient      & 0.01           \\
Valid Speech loss coefficient   & 40             \\
Speech entropy coefficient      & 0.04           \\
Batch size                      & 32             \\
Gradient clip                   & 1.0            \\
Steps per episode               & 100            \\ \hline
Policy Networks (Actors)        &                \\ \hline
Num. Layers                     & 3              \\
Feedforward network hidden size & 768            \\
GRU hidden size                 & 768            \\ \hline
\multicolumn{2}{l}{Value Predictor (Critic)}     \\ \hline
Num. Layers                     & 2              \\
Feedforward network hidden size & 768            \\ \hline
\multicolumn{2}{l}{Appended Encoder}             \\ \hline
Num. layers                     & 3              \\
Num. attention heads            & 3              \\
Feedforward network hidden size & 768    
\end{tabular}
\caption{RL experiments hyperparameters. All pre-training encoder hyperparameters are as found earlier in Table~\ref{tab:suphyper}.}
\label{tab:rlhyper}
\end{table}

\subsubsection{Learning Curves} 
\begin{figure*}[!h]
    \centering
    \begin{subfigure}{0.32\textwidth}
    \centering
    \includegraphics[width=\linewidth]{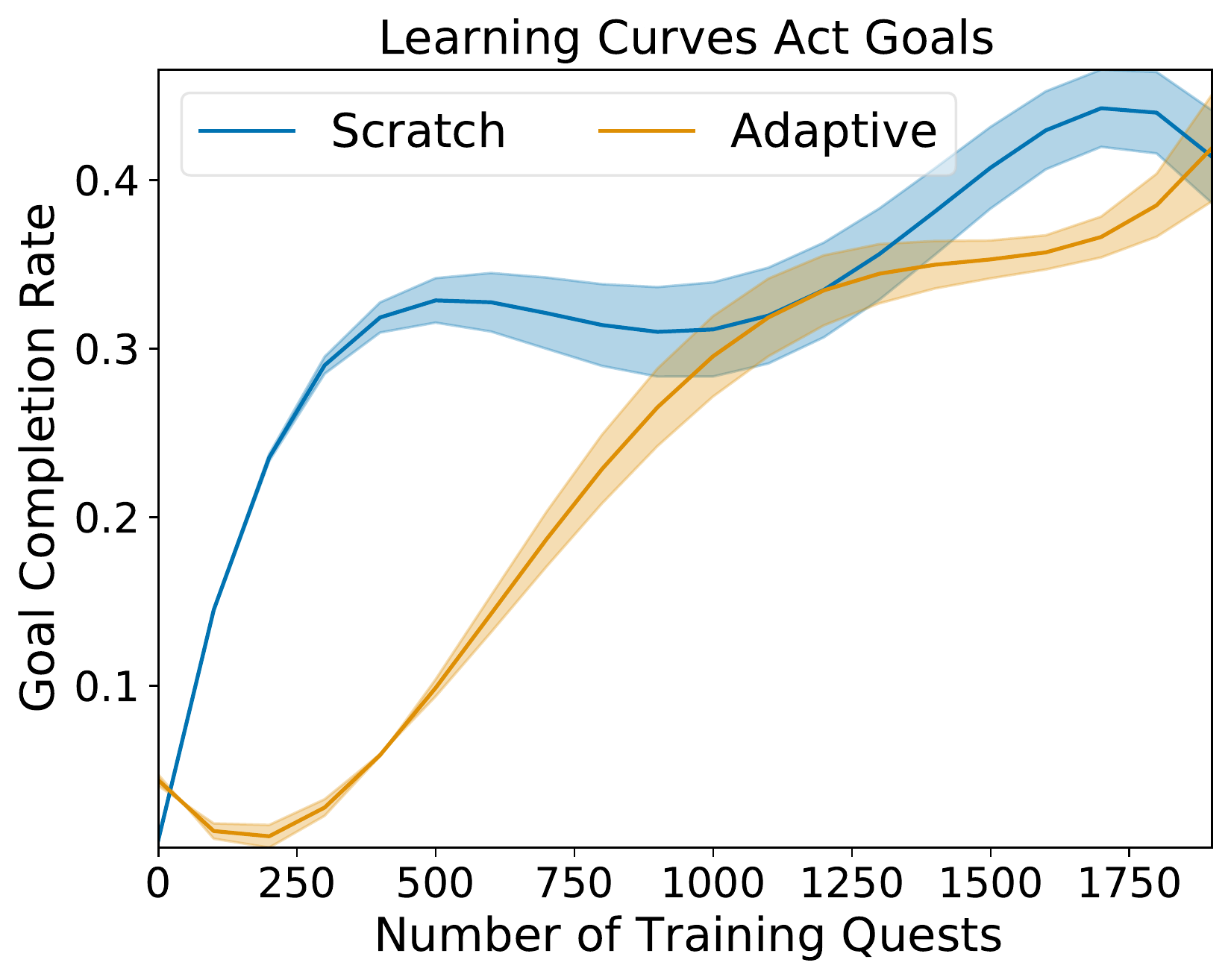}
    \caption{Act Goals}
    \label{fig:lcrlact}
    \end{subfigure}
    \begin{subfigure}{0.33\textwidth}
    \centering
    \includegraphics[width=\linewidth]{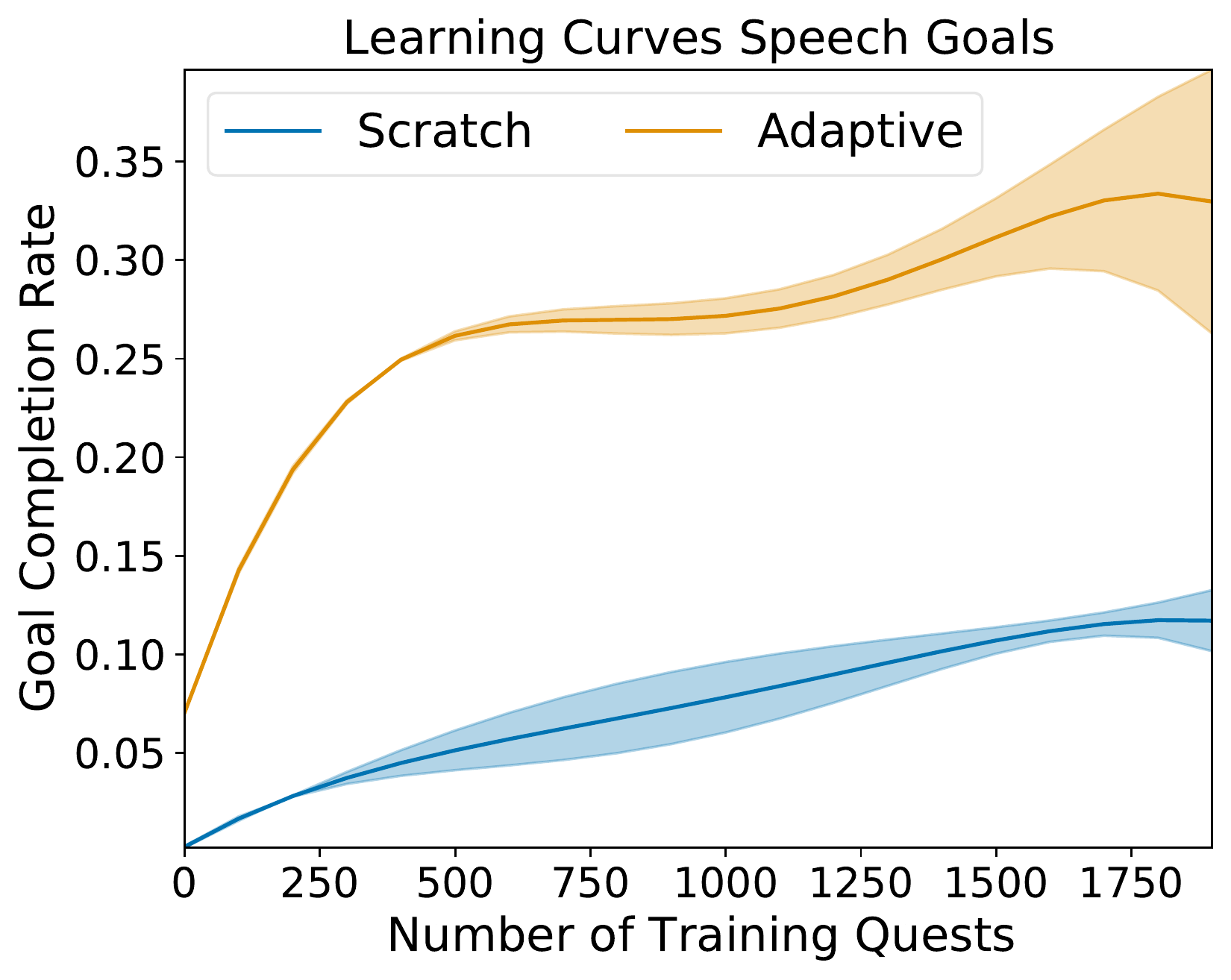}
    \caption{Speech Goals}
    \label{fig:lcrlspeech}
    \end{subfigure}
    \begin{subfigure}{0.32\textwidth}
    \centering
    \includegraphics[width=\linewidth]{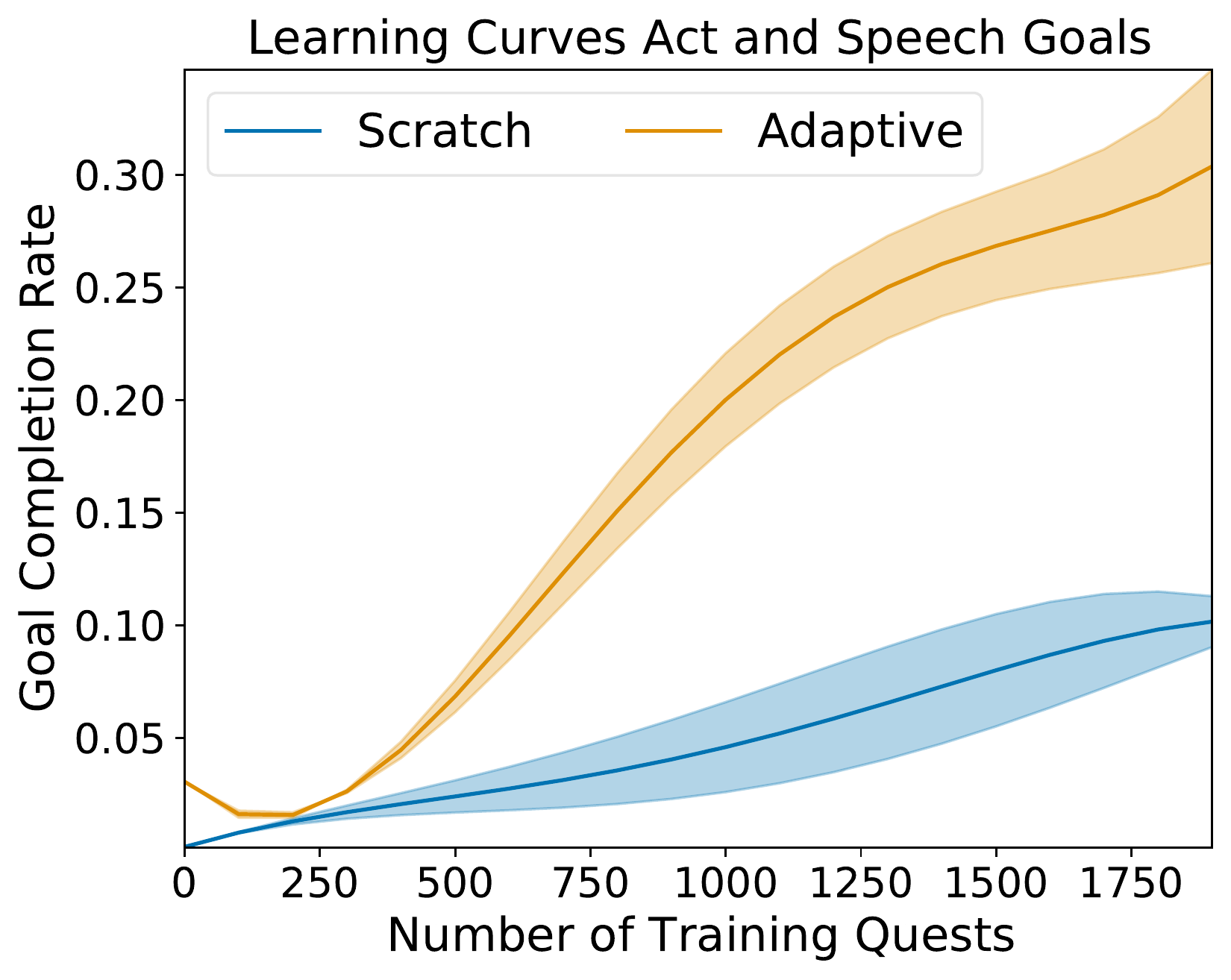}
    \caption{Act and Speech Goals}
    \label{fig:lcrlall}
    \end{subfigure}
    \caption{Encoder Types RL Learning Curves averaged over 3 independent runs.}
    \label{fig:lcrl}
\end{figure*}

The first set of results, seen in Figure~\ref{fig:lcrl} shows that both Scratch and Adaptive models gain performance across the board in terms of their ability to act and speak given more training quests.
Unlike the supervised tasks, the Scratch model generally benefits less than the Adaptive model from having more data.

\subsubsection{Switch Type Ablations} 
\begin{figure*}[!h] 
    \centering
    \begin{subfigure}{0.34\textwidth}
    \centering
    \includegraphics[width=\linewidth]{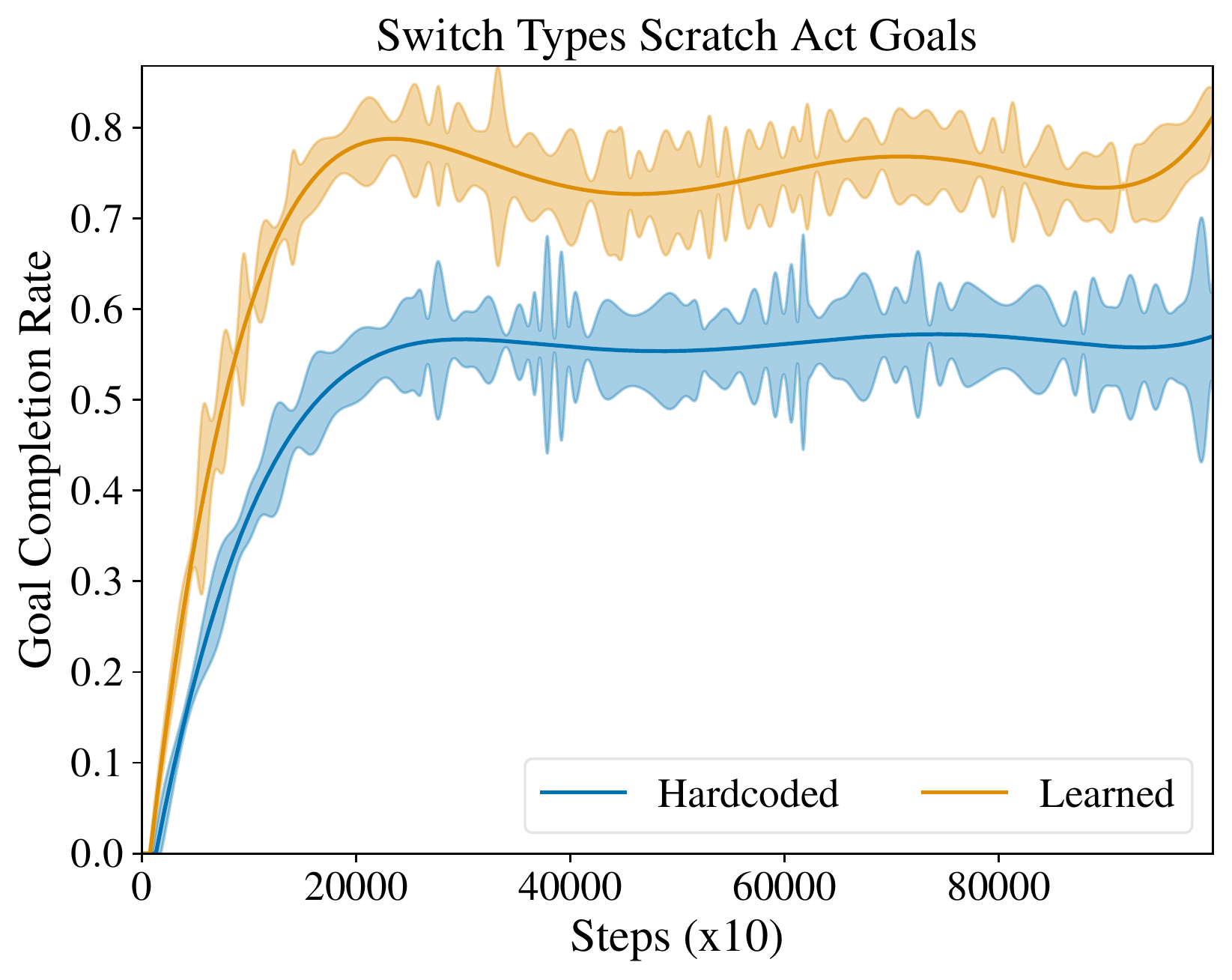}
    \caption{Act Goals}
    \label{fig:priorityscratchact}
    \end{subfigure}
    \begin{subfigure}{0.34\textwidth}
    \centering
    \includegraphics[width=\linewidth]{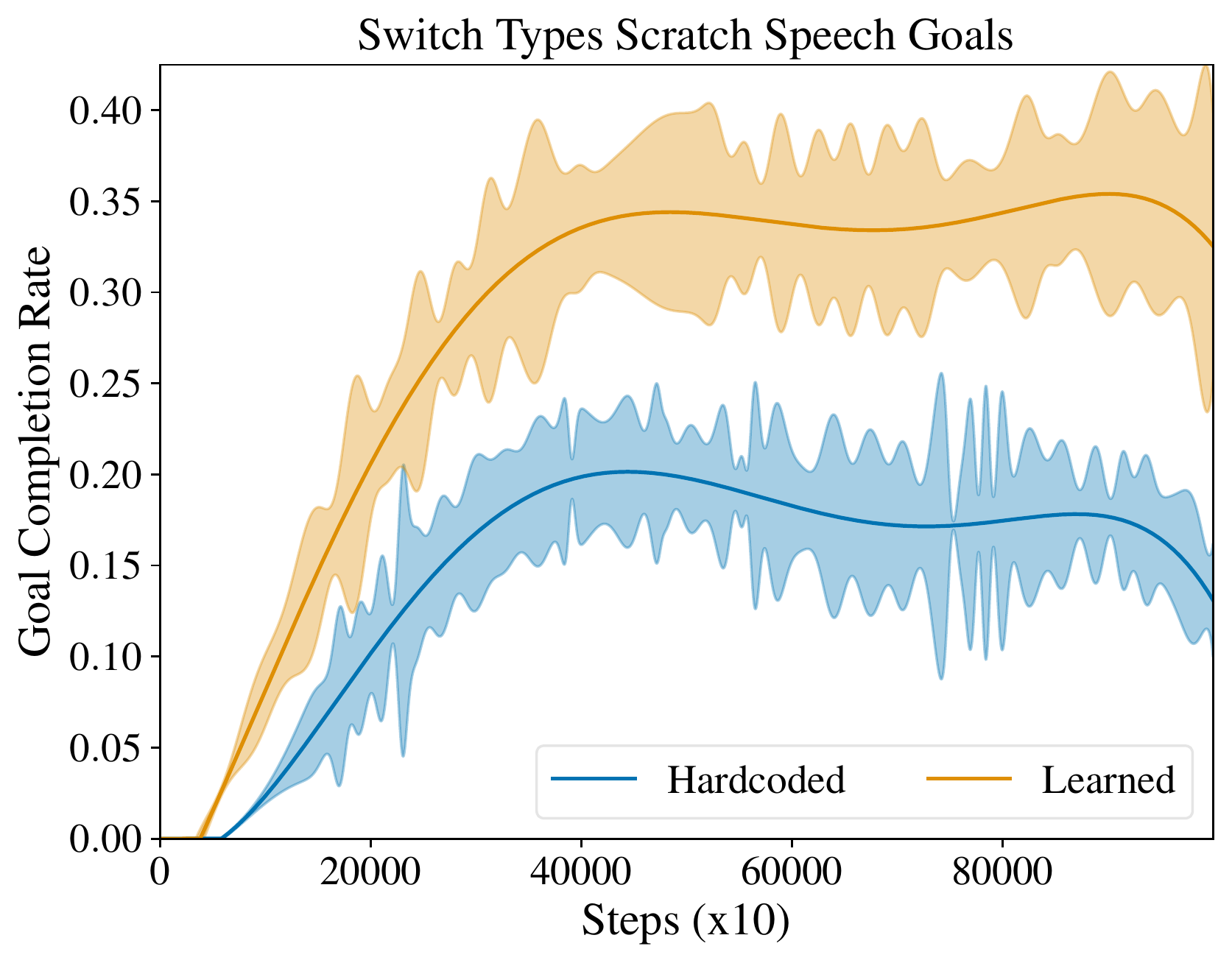}
    \caption{Speech Goals}
    \label{fig:priorityscratchspeech}
    \end{subfigure}
    \caption{Switch Types Reward Curves for the Scratch Model averaged over 3 independent runs.}
    \label{fig:switchscratch}
\end{figure*}

\begin{figure*}[!h] 
    \centering
    \begin{subfigure}{0.34\textwidth}
    \centering
    \includegraphics[width=\linewidth]{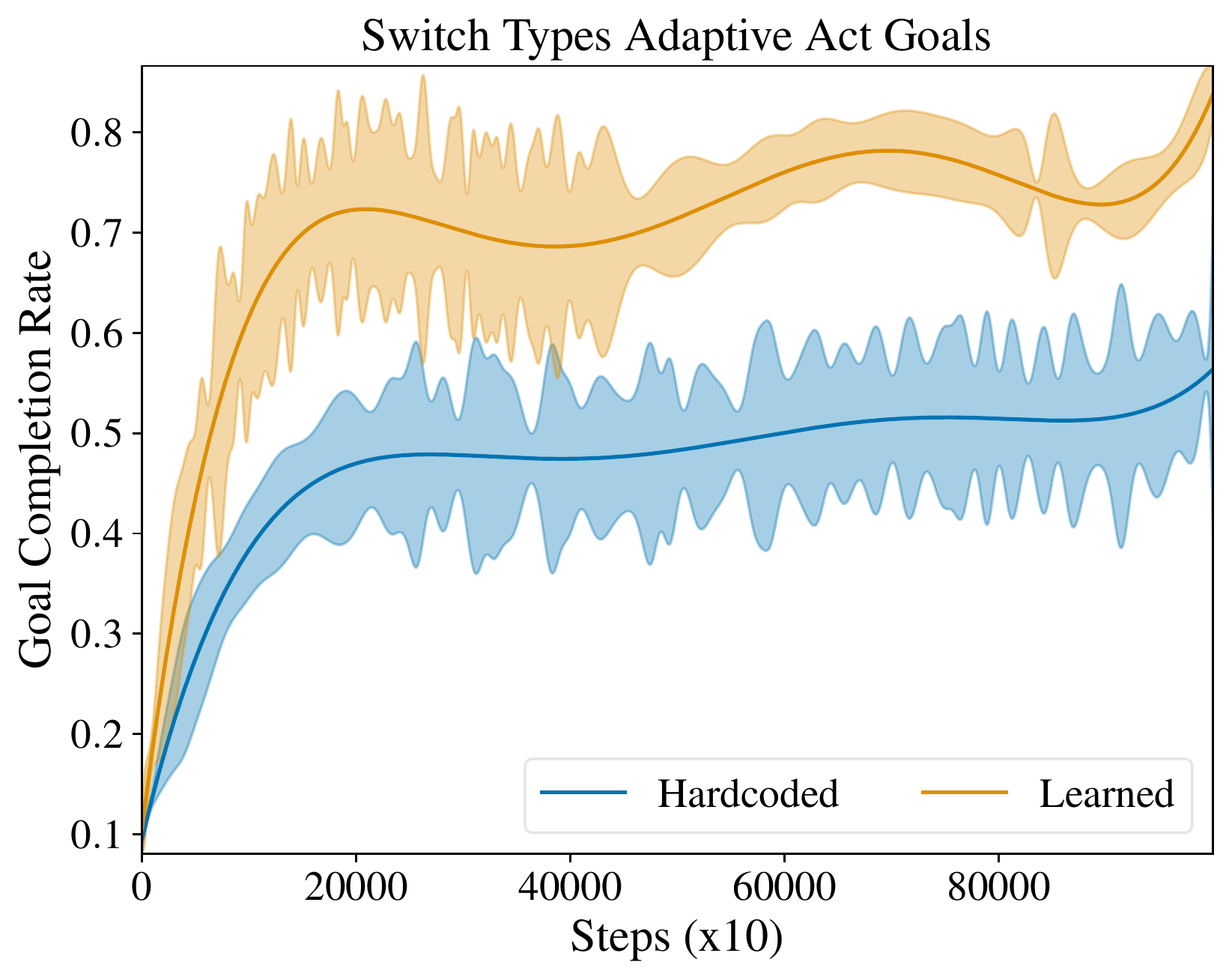}
    \caption{Act Goals}
    \label{fig:prioritysinkact}
    \end{subfigure}
    \begin{subfigure}{0.34\textwidth}
    \centering
    \includegraphics[width=\linewidth]{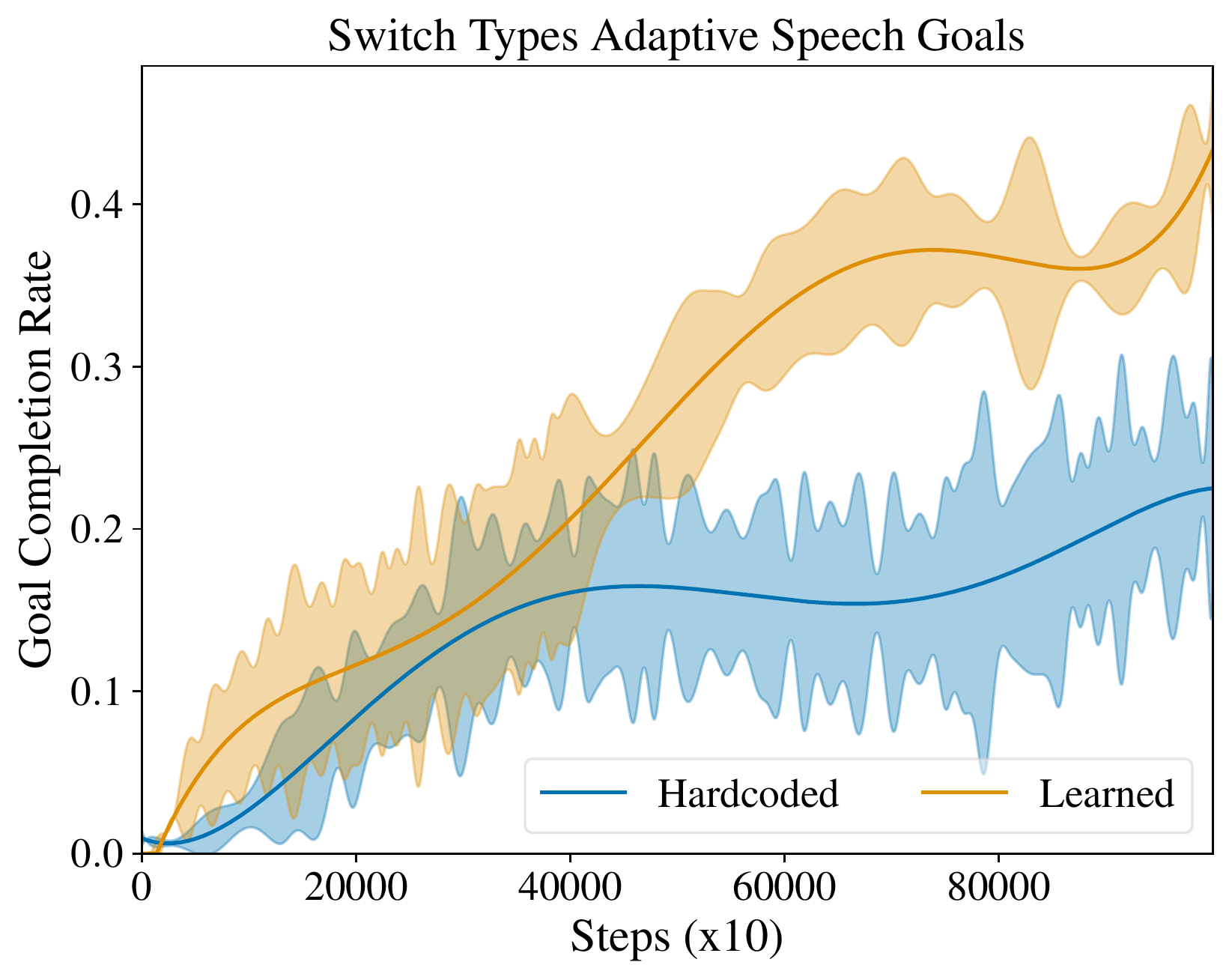}
    \caption{Speech Goals}
    \label{fig:prioritysinkspeech}
    \end{subfigure}
    \caption{Switch Types Reward Curves for the Adaptive Model averaged over 3 independent runs.}
    \label{fig:switchsink}
\end{figure*}

The second set of results involve ablating having a learned switch that uses the input training data and a hardcoded switch-
The learned switch is as described in Section~\ref{sec:lightrl}: it outputs an action every $k$ dialogue utterances; where during training $k$ is chosen to match the ratio of utterances to actions on that particular quest from the human demonstrations, and during testing, $k$ is chosen to match the average action to utterance ratio.
The hardcoded switch is where the agent outputs an action chosen every $N$ steps across all quests---here $N=3$ is the chosen hyperparameter.
Table~\ref{tab:switch} shows that having a learned switch increases zero-shot generalization performance and Figures~\ref{fig:switchscratch},~\ref{fig:switchsink} show that having a learned switch improves sample efficiency by enabling the LIGHT agent to reach asymptotic performance in fewer steps in both the Scratch and Adaptive models.
\begin{table}[!h]
\centering
\scriptsize
\begin{tabular}{|l|l|l|l|l|}
\hline
\multicolumn{1}{|c|}{\multirow{2}{*}{\textbf{Model}}} & \multicolumn{3}{c|}{\textbf{Reinforcement Learning}}                                                                                       \\ 
\multicolumn{1}{|c|}{}                                & \multicolumn{1}{c|}{\textbf{Act Goals}} & \multicolumn{1}{c|}{\textbf{Speech Goals}} & \multicolumn{1}{c|}{\textbf{Act \& Speech Goals}} \\ \hline \hline
Scratch                                               & 0.418                           & 0.118                               & 0.103                                    \\
\multicolumn{1}{|r|}{Hardcoded Switch}           & 0.295                             & 0.0947                                 & 0.066                                                                                      \\ \hline \hline

Adaptive                                                  & \textbf{0.42}                              & \textbf{0.33}                                 & \textbf{0.303}                                       \\
\multicolumn{1}{|r|}{Hardcoded Switch}           & 0.274                             & 0.294                                 & 0.236                                            \\ 
\hline   
\end{tabular}
\caption{Encoder Type RL Zero-Shot Evaluations averaged over 3 independent runs. Act goals and speech goals are as described in Section~\ref{sec:env}. Standard deviations for all experiments are less than 0.01. The ``Act \& Speech Goals'' column refers to quests where the agent has simultaneously achieved both types of goals within the allotted one episode.}
\label{tab:switch}
\end{table}

\clearpage
\subsubsection{Self Act Completion Transcripts}
We pick 3 transcripts of the LIGHT agent playing the quests.
Each of these transcripts are from the Adaptive model on the test set of human demonstrations during zero-shot evaluations.
We pick samples where the agent achieves the {\em act goal} itself and also simultaneously achieves the {\em speech goal}.
The blue and right aligned text is the LIGHT agent trained with RL and the gray and left aligned text is the partner agent.

\begin{figure}[!h]
    \centering
    \scriptsize
\begin{minipage}{\textwidth}
\begin{tabularx}{\textwidth}{|r|X|}
\hline
\multicolumn{1}{|l|}{\textbf{Setting}}   & The fishing store is a small one room stone building with wares laid about on tables. One can see fishing poles, wooden buckets with dirt and bait inside, along with some mounted trophies, and a skeleton.  \\ \hline
\rowcolor[HTML]{E0E0E0}
\multicolumn{1}{|l|}{\cellcolor[HTML]{E0E0E0}{\textbf{Partner:}}}   & Fish seller.   \\
\rowcolor[HTML]{E0E0E0}
Persona & I sell fish in the kingdom. I usually perch on the high seas. I catch thousands of fish each year.
  \\
\rowcolor[HTML]{E0E0E0}
Carrying &  fish \\ \hline
\rowcolor[HTML]{D5EBFC}
\multicolumn{1}{|l|}{\cellcolor[HTML]{D5EBFC}{\textbf{Self:}}} &  Mermaid hunter.\\
\rowcolor[HTML]{D5EBFC}
Persona & I am a fierce hunter. I bring down mermaids. I spend most of my time on the high seas.\\
\rowcolor[HTML]{D5EBFC}
Carrying & crossbow, skeleton, coins  \\
\rowcolor[HTML]{D5EBFC}
Motivation & I will hit the fish seller to remind them who I am.
\\
\hline
\end{tabularx}

\end{minipage}
\begin{elaboration}
  \parbox{.99\textwidth}{
 I'll get them!!
}
\end{elaboration}
\begin{elaborationr}
  \parbox{.99\textwidth}{
\hspace*{\fill} {\em follow fish seller}
}
\end{elaborationr}
\begin{elaboration}
  \parbox{.99\textwidth}{
{\em hit mermaid hunter}
}
\end{elaboration}
\begin{elaborationr}
  \parbox{.99\textwidth}{
\hspace*{\fill} That's it! I am going to eat you. Say your last words.
}
\end{elaborationr}
\begin{elaboration}
  \parbox{.99\textwidth}{
{\em hug mermaid hunter}
}
\end{elaboration}
\begin{elaborationr}
  \parbox{.99\textwidth}{
\hspace*{\fill} Do you see these skeletons I am carrying? They are mermaids I’ve already devoured!
}
\end{elaborationr}
\begin{elaboration}
  \parbox{.99\textwidth}{
{\em get skeleton}
}
\end{elaboration}
\begin{elaborationr}
  \parbox{.99\textwidth}{
\hspace*{\fill} Get away from my skeletons! The bones are my trophy!
}
\end{elaborationr}
\begin{elaboration}
  \parbox{.99\textwidth}{
{\em give skeleton to mermaid hunter}}
\end{elaboration}
\begin{elaborationr}
  \parbox{.99\textwidth}{
\hspace*{\fill}It is all your fault my last mermaid got away!
}
\end{elaborationr}
\begin{elaboration}
  \parbox{.99\textwidth}{
And you are a simpleton, just take this fish. I do not even care about payment.}
\end{elaboration}
\begin{elaborationr}
  \parbox{.99\textwidth}{
\hspace*{\fill} I DONT WANT YOUR DAMN FISH
}
\end{elaborationr}
\begin{elaboration}
  \parbox{.99\textwidth}{
{\em get coins from fist}}
\end{elaboration}
\begin{elaborationr}
  \parbox{.99\textwidth}{
\hspace*{\fill}{\em hit fish seller}
}
\end{elaborationr}
\end{figure}

\begin{figure}[!h]
    \centering
    \scriptsize
\begin{minipage}{\textwidth}
\begin{tabularx}{\textwidth}{|r|X|}
\hline
\multicolumn{1}{|l|}{\textbf{Setting}}  & The king's bedroom. The walls are tall and stone. They are coated with colorful tapestries showing the kings of years past. A large stone fireplace across from the bed that is large enough to keep the king warm even on the coldest nights. A double thick wooden door with a large lock on each side of the room.
 \\ \hline
\rowcolor[HTML]{E0E0E0}
\multicolumn{1}{|l|}{\cellcolor[HTML]{E0E0E0}{\textbf{Partner:}}}   & Royal dog.   \\
\rowcolor[HTML]{E0E0E0}
Persona & I am the royal dog, fat, incontinent, and lazy. I eat off my own porcelain plate, when I am not hand fed by the queen. I pee where I like. I stole food from the poor when I was young, but I cannot waddle up to them fast enough these days. I sleep between the royal couple.
 \\
\rowcolor[HTML]{E0E0E0}
Carrying & Nothing. \\ \hline
\rowcolor[HTML]{D5EBFC}
\multicolumn{1}{|l|}{\cellcolor[HTML]{D5EBFC}{\textbf{Self:}}} &  Queen.\\
\rowcolor[HTML]{D5EBFC}
Persona & I was the daughter of a high ranking nobleman overseas. To make a trade alliance with the King, my parents offered me in marriage. It wasn't my idea, but it has turned out very well. I've produced two living sons as heirs, and the king treats me kindly. I spend my time doing embroidery and talking with my ladies in waiting.
\\
\rowcolor[HTML]{D5EBFC}
Carrying & Nothing. \\
\rowcolor[HTML]{D5EBFC}
Motivation & I want to get to the large stone fireplace.\\
\hline
\end{tabularx}

\end{minipage}
\begin{elaborationr}
  \parbox{.99\textwidth}{
\hspace*{\fill} I am having turbulent thoughts regarding my faith and own morality.
}
\end{elaborationr}
\begin{elaboration}
  \parbox{.99\textwidth}{
I want to jump into the fireplace.
}
\end{elaboration}
\begin{elaborationr}
  \parbox{.99\textwidth}{
\hspace*{\fill} Don't, lets go to the fireplace and sit.
}
\end{elaborationr}
\begin{elaboration}
  \parbox{.99\textwidth}{
I go to fireplace.
}
\end{elaboration}
\begin{elaborationr}
  \parbox{.99\textwidth}{
\hspace*{\fill}{\em get large stone fireplace}}
\end{elaborationr}
\end{figure}

\clearpage
\begin{figure}[!h]
    \centering
    \scriptsize
\begin{minipage}{\textwidth}
\begin{tabularx}{\textwidth}{|r|X|}
\hline
\multicolumn{1}{|l|}{\textbf{Setting}}   & The sheriff's office is a small building on the southern edge of the farm. There is a desk with decrees from the king and other papers strewn about. Hanging on the wall are swords and other weapons. The room is small and cluttered, with various books and other things strewn about. There is a guard here.
 \\ \hline
\rowcolor[HTML]{E0E0E0}
\multicolumn{1}{|l|}{\cellcolor[HTML]{E0E0E0}{\textbf{Partner:}}}   & Lawyer.   \\
\rowcolor[HTML]{E0E0E0}
Persona & I am a very good lawyer. I always win in my cases. I would like to work as a district attorney.
 \\
\rowcolor[HTML]{E0E0E0}
Carrying & Nothing. \\ \hline
\rowcolor[HTML]{D5EBFC}
\multicolumn{1}{|l|}{\cellcolor[HTML]{D5EBFC}{\textbf{Self:}}} &  Sheriff.\\
\rowcolor[HTML]{D5EBFC}
Persona & I keep our town safe. I make sure to check the caravans for bandits. If the King needs anything extra hands in the kingdom he can count on me.
 \\
\rowcolor[HTML]{D5EBFC}
Carrying & Nothing.  \\
\rowcolor[HTML]{D5EBFC}
Motivation & I want to remove the shield and hit the lawyer with it. \\
\hline
\end{tabularx}
 
\end{minipage}
\begin{elaborationr}
  \parbox{.99\textwidth}{
\hspace*{\fill} Next time I'll hit you with this shield! Tell the truth, it was you that went after the king! I don't buy the frame story. No matter how good a lawyer you are. You aren't fooling me!
}
\end{elaborationr}
\begin{elaboration}
  \parbox{.99\textwidth}{
{\em get shield}
}
\end{elaboration}
\begin{elaborationr}
  \parbox{.99\textwidth}{
\hspace*{\fill}{\em remove knife}
}
\end{elaborationr}
\begin{elaboration}
  \parbox{.99\textwidth}{
Hahahahaha. I'm off, don't like to fight! You are a fool sheriff. I can run faster than you can ever imagine.
}
\end{elaboration}
\begin{elaborationr}
  \parbox{.99\textwidth}{
\hspace*{\fill} {\em follow lawyer}
}
\end{elaborationr}
\begin{elaboration}
  \parbox{.99\textwidth}{
This is my shield now! Goodbye! I'm off.}
\end{elaboration}
\begin{elaborationr}
  \parbox{.99\textwidth}{
\hspace*{\fill}Stop!!
}
\end{elaborationr}
\begin{elaboration}
  \parbox{.99\textwidth}{
{\em put shield on large and well stocked wine rack}}
\end{elaboration}
\begin{elaborationr}
  \parbox{.99\textwidth}{
\hspace*{\fill}{\em get shield from large and well stocked wine rack}
}
\end{elaborationr}
\begin{elaboration}
  \parbox{.99\textwidth}{
{\em hug sheriff}}
\end{elaboration}
\begin{elaborationr}
  \parbox{.99\textwidth}{
\hspace*{\fill}{\em hit lawyer}
}
\end{elaborationr}
\begin{elaboration}
  \parbox{.99\textwidth}{
Goodbye.}
\end{elaboration}
\begin{elaborationr}
  \parbox{.99\textwidth}{
\hspace*{\fill}Alright, lawyer. You may go now.
}
\end{elaborationr}
\end{figure}

\subsubsection{Partner Act Completion Transcripts}
Similarly to the previous section, these transcripts are from the Adaptive model on the test set of human demonstrations during zero-shot evaluations.
We pick samples where the {\em act goal} is achieved via the partner and the agent simultaneously achieves the {\em speech goal} itself.
The blue and right aligned text is the LIGHT agent trained with RL and the gray and left aligned text is the partner agent.

\begin{figure}[!h]
    \centering
    \scriptsize
\begin{minipage}{\textwidth}
\begin{tabularx}{\textwidth}{|r|X|}
\hline
\multicolumn{1}{|l|}{\textbf{Setting}}   & Up at the top of the tower, among the gray clouds, is the prison room. The room is round, with a circumference of about 15 feet. The walls are barren and made of stone; only one barred window gives a view out onto the vast landscape. A small bed and a dirty wooden bucket are the only decorations. The desolate fairy spends day and night laying on that small bed, wondering where things went wrong.
 \\ \hline
\rowcolor[HTML]{E0E0E0}
\multicolumn{1}{|l|}{\cellcolor[HTML]{E0E0E0}{\textbf{Partner:}}}   & Fairy.   \\
\rowcolor[HTML]{E0E0E0}
Persona & I'm a fairy that lives in the forest. I like to fly around with my siblings and play games. I try to hide from humans. If they find me, I'll grant their wishes.
 \\
\rowcolor[HTML]{E0E0E0}
Carrying & Nothing. \\ \hline
\rowcolor[HTML]{D5EBFC}
\multicolumn{1}{|l|}{\cellcolor[HTML]{D5EBFC}{\textbf{Self:}}} &  Witch.\\
\rowcolor[HTML]{D5EBFC}
Persona & Halloween is my favorite time of the year.  My familiar is a cat named Helix.  I have a spell book that is 300 years old.  I search the forest for herbs and berries for my potions.
 \\
\rowcolor[HTML]{D5EBFC}
Carrying & Nothing.  \\
\rowcolor[HTML]{D5EBFC}
Motivation & I need the dirty wooden bucket to make a potion.\\
\hline
\end{tabularx}

\end{minipage}
\begin{elaboration}
  \parbox{.99\textwidth}{
{\em Take the doll}
}
\end{elaboration}
\begin{elaborationr}
  \parbox{.99\textwidth}{
\hspace*{\fill} {\em hug fairy}
}
\end{elaborationr}
\begin{elaboration}
  \parbox{.99\textwidth}{
I will take this doll back to the palace with me.
}
\end{elaboration}
\begin{elaborationr}
  \parbox{.99\textwidth}{
\hspace*{\fill} You do you fairy.
}
\end{elaborationr}

\begin{elaboration}
  \parbox{.99\textwidth}{
{\em hug witch}
}
\end{elaboration}
\begin{elaborationr}
  \parbox{.99\textwidth}{
\hspace*{\fill} You know whats going on w/ the princess?
}
\end{elaborationr}
\begin{elaboration}
  \parbox{.99\textwidth}{
She just disappeared suddenly late at night without a sound heard
}
\end{elaboration}
\begin{elaborationr}
  \parbox{.99\textwidth}{
\hspace*{\fill} Anyway, I will make this potion and drink it, but I do not think it will help me.
}
\end{elaborationr}
\begin{elaboration}
  \parbox{.99\textwidth}{
I will drink the potion but why make it.
}
\end{elaboration}
\begin{elaborationr}
  \parbox{.99\textwidth}{
\hspace*{\fill} Because we're in a prison tower. Duh. Also give me that bucket or I'll feed you to my cat!
}
\end{elaborationr}
\begin{elaboration}
  \parbox{.99\textwidth}{
{\em give dirty wooden bucket to witch}}
\end{elaboration}
\end{figure}

\begin{figure*}[!h]
    \centering
    \scriptsize
\begin{minipage}{\textwidth}
\begin{tabularx}{\textwidth}{|r|X|}
\hline
\multicolumn{1}{|l|}{\textbf{Setting}}   & The throne room. Very luxurious, with details of 12 carat gold and diamonds with a very comfortable cushions where the king usually takes a break.
\\ \hline
\rowcolor[HTML]{E0E0E0}
\multicolumn{1}{|l|}{\cellcolor[HTML]{E0E0E0}{\textbf{Partner:}}}   & Powerful but aged wizard.   \\
\rowcolor[HTML]{E0E0E0}
Persona & I am the official wizard to the king. I give him advice on how to run the kingdom. I enjoy and love using magic in my everyday life. I'm getting old now but I still have a zest for life. I really enjoy doing magic for the local peasants in the village.
 \\
\rowcolor[HTML]{E0E0E0}
Wielding & robe, staff \\ \hline
\rowcolor[HTML]{D5EBFC}
\multicolumn{1}{|l|}{\cellcolor[HTML]{D5EBFC}{\textbf{Self:}}} &  King's trusted adviser.\\
\rowcolor[HTML]{D5EBFC}
Persona & I am the king's trusted adviser. I hold the discretion of the king. I guard the secrets and strategies that make my kingdom thrive. I live a lonely life because others cannot be trusted.
 \\
\rowcolor[HTML]{D5EBFC}
Wielding & slippers, robes  \\
\rowcolor[HTML]{D5EBFC}
Motivation & I need to bring the king's pet to him\\
\hline
\end{tabularx}

\end{minipage}
\begin{elaborationr}
  \parbox{.99\textwidth}{
\hspace*{\fill} Well, I was doing what you told me to do. I was going to eat him but there was hardly any meat on him.
}
\end{elaborationr}
\begin{elaboration}
  \parbox{.99\textwidth}{
I killed him before he could grow.
}
\end{elaboration}
\begin{elaborationr}
  \parbox{.99\textwidth}{
\hspace*{\fill} Carry me to pond.
}
\end{elaborationr}
\begin{elaboration}
  \parbox{.99\textwidth}{
Yes, milord, I shall do as you command.
}
\end{elaboration}
\begin{elaborationr}
  \parbox{.99\textwidth}{
\hspace*{\fill} {\em remove slippers}
}
\end{elaborationr}
\begin{elaboration}
  \parbox{.99\textwidth}{
Where is King Fulmer's pet?
}
\end{elaboration}
\begin{elaborationr}
  \parbox{.99\textwidth}{
\hspace*{\fill} Is he not in his bed?
}
\end{elaborationr}
\begin{elaboration}
  \parbox{.99\textwidth}{
Yes, I guess you need to bring him in.}
\end{elaboration}
\begin{elaborationr}
  \parbox{.99\textwidth}{
\hspace*{\fill}{\em follow powerful but aged wizard}
}
\end{elaborationr}
\begin{elaboration}
  \parbox{.99\textwidth}{
I want to become the new head wizard.}
\end{elaboration}
\begin{elaborationr}
  \parbox{.99\textwidth}{
\hspace*{\fill}Nope he just probably throw you overboard.
}
\end{elaborationr}
\begin{elaboration}
  \parbox{.99\textwidth}{
The wizard is evil? Dang it. Help me.}
\end{elaboration}
\begin{elaborationr}
  \parbox{.99\textwidth}{
\hspace*{\fill}{You're not the lion, I need to get the lion out.}
}
\end{elaborationr}
\begin{elaboration}
  \parbox{.99\textwidth}{
{\em give large pet lion to king's trusted adviser}}
\end{elaboration}
\end{figure*}

\begin{figure*}[!h]
    \centering
    \scriptsize
\begin{minipage}{\textwidth}
\begin{tabularx}{\textwidth}{|r|X|}
\hline
\multicolumn{1}{|l|}{\textbf{Setting}}  & The Tent. Shabby, torn fabric, stained by weeks of travel and bleached from the harsh sun. Long, thin sticks held the
fabric up so it formed a triangle with the earth. The sticks were buried deep within the shifting orange sand to hold off
the blistering wind.    \\ \hline
\rowcolor[HTML]{E0E0E0}
\multicolumn{1}{|l|}{\cellcolor[HTML]{E0E0E0}{\textbf{Partner:}}}   & Person.   \\
\rowcolor[HTML]{E0E0E0}
Persona & I'm just a guy. I work at the saw mill every day. I'm a good villager. \\
\rowcolor[HTML]{E0E0E0}
Carrying & Nothing. \\ \hline
\rowcolor[HTML]{D5EBFC}
\multicolumn{1}{|l|}{\cellcolor[HTML]{D5EBFC}{\textbf{Self:}}} &  Military Commander.\\
\rowcolor[HTML]{D5EBFC}
Persona & I am the military commander of the village. It is my duty to train and lead our soldiers into battle. I am a stern commander and I expect the best from my troops. \\
\rowcolor[HTML]{D5EBFC}
Carrying & sword  \\
\rowcolor[HTML]{D5EBFC}
Motivation & I need to get sand to use as a tactical planning prop.\\
\hline
\end{tabularx}

\end{minipage}

\begin{elaboration}
  \parbox{.99\textwidth}{
Just ensure the home front is properly protected.
}
\end{elaboration}
\begin{elaborationr}
  \parbox{.99\textwidth}{
\hspace*{\fill} {\em hug person}
}
\end{elaborationr}
\begin{elaboration}
  \parbox{.99\textwidth}{
Perfect! God speed commander.

}
\end{elaboration}
\begin{elaborationr}
  \parbox{.99\textwidth}{
\hspace*{\fill} Yes. I need to prepare.
}
\end{elaborationr}
\begin{elaboration}
  \parbox{.99\textwidth}{
Now. I heard the enemy is coming.}
\end{elaboration}
\begin{elaborationr}
  \parbox{.99\textwidth}{
\hspace*{\fill} {\em pick up stick}
}
\end{elaborationr}
\begin{elaboration}
  \parbox{.99\textwidth}{
{\em get stick}}
\end{elaboration}
\begin{elaborationr}
  \parbox{.99\textwidth}{
\hspace*{\fill} Where's the sand?
}
\end{elaborationr}
\begin{elaboration}
  \parbox{.99\textwidth}{
{\em give sand to military commander}}
\end{elaboration}
\end{figure*}

\end{document}